\documentclass[10pt,twocolumn,letterpaper]{article}

\usepackage{cvpr}
\usepackage{hyperref}
\usepackage{url}
\usepackage[svgnames]{xcolor}
\usepackage{tcolorbox}
\tcbuselibrary{skins, breakable}
\usepackage{multirow}
\usepackage{tabularx}
\usepackage{array}
\usepackage{subcaption}
\usepackage{wrapfig}
\usepackage{xcolor}
\usepackage[utf8]{inputenc} 
\usepackage[T1]{fontenc}    
\usepackage{url}            
\usepackage{booktabs}       
\usepackage{amsfonts}       
\usepackage{nicefrac}       
\usepackage{microtype}      
\usepackage{xcolor}         
\usepackage{caption}
\usepackage{float}
\usepackage{url}            
\usepackage{booktabs}       
\usepackage{amsfonts}       
\usepackage{nicefrac}       
\usepackage{microtype}      
\usepackage{lipsum}		
\usepackage{graphicx}
\usepackage{natbib}
\usepackage{doi}
\usepackage{times}
\usepackage{multirow}
\usepackage[table]{xcolor} 
\usepackage{soul}
\usepackage{bbding}
\usepackage{url}
\usepackage{amsmath}
\usepackage{amsthm}
\usepackage{algorithm}
\usepackage{algorithmic}
\usepackage{newfloat}
\usepackage{bbding}
\usepackage{listings}
\usepackage{tikz}
\usepackage{comment}
\usepackage{amssymb}
\usepackage{color}
\usepackage[utf8]{inputenc}
\usepackage[T1]{fontenc}
\usepackage{xcolor}

\usepackage{hyperref}
\hypersetup{
    colorlinks=true,
    citebordercolor=blue,
    linkbordercolor=red,
    urlbordercolor=magenta
}

\definecolor{cvprblue}{rgb}{0.21,0.49,0.74}
\def\paperID{xxxx}
\def\confName{CVPR}
\def\confYear{2026}

\title{PersonaVLM: Long-Term Personalized Multimodal LLMs}
\author{
    Chang Nie$^1$ \quad 
    Chaoyou Fu$^{1}$\thanks{Corresponding author.} \quad 
    Yifan Zhang$^2$ \quad 
    Haihua Yang$^{2}$\thanks{Project leader.} \quad 
    Caifeng Shan$^1$ \vspace{2mm}\\
    $^1$Nanjing University \quad $^2$ByteDance \\
    {\tt\small changnie@smail.nju.edu.cn, bradyfu24@gmail.com}
}
\begin{document}
\maketitle
\begin{abstract}
Multimodal Large Language Models (MLLMs) serve as daily assistants for millions. However, their ability to generate responses aligned with individual preferences remains limited.
Prior approaches enable only static, single-turn personalization through input augmentation or output alignment, and thus fail to capture users’ evolving preferences and personality over time (see Fig.~\ref{img1}).
In this paper, we introduce \textbf{PersonaVLM}, an innovative personalized multimodal agent framework designed for long-term personalization.
It transforms a general-purpose MLLM into a personalized assistant by integrating three key capabilities:
(a) \textbf{\textit{Remembering}}: It proactively extracts and summarizes chronological multimodal memories from interactions, consolidating them into a personalized database.
(b) \textbf{\textit{Reasoning}}: It conducts multi-turn reasoning by retrieving and integrating relevant memories from the database. 
(c) \textbf{\textit{Response Alignment}}: It infers the user's evolving personality throughout long-term interactions to ensure outputs remain aligned with their unique characteristics.
For evaluation, we establish \textbf{Persona-MME}, a comprehensive benchmark comprising over 2,000 curated interaction cases, designed to assess long-term MLLM personalization across seven key aspects and 14 fine-grained tasks.
Extensive experiments validate our method's effectiveness, improving the baseline by 22.4\% (Persona-MME) and 9.8\% (PERSONAMEM) under a 128$k$ context, while outperforming GPT-4o by 5.2\% and 2.0\%, respectively.
Project page:
\url{https://PersonaVLM.github.io}.

\end{abstract}

\section{Introduction}
\label{sec:intro}

Multimodal Large Language Models (MLLMs) are increasingly integrated into the daily lives of millions of users~\cite{achiam2023gpt,yao2024minicpm}, serving as assistants, creative partners, and companions~\cite{li2024multimodal,xu2024can,yin2024survey}.
As their adoption grows, user expectations are shifting from general-purpose problem-solving towards personalized and empathetic long-term experiences~\cite{li2024hello,wu2024personalized}.
This shift poses a critical question:
\textit{\textbf{How can we evolve a general MLLM into a truly personalized assistant that accurately infers user intent, dynamically aligns its behavior with individual preferences and personality, and persistently remembers user-specific multimodal information over time?}}
Addressing this question not only enhances user satisfaction and trust but also unlocks the significant value of MLLMs in domains like recommendation~\cite{wang2024towards}, healthcare~\cite{alsaad2024multimodal}, and education~\cite{yu2024mooc}, to name a few.

\begin{figure*}[t]
\centering
\includegraphics[width=1.0\textwidth]{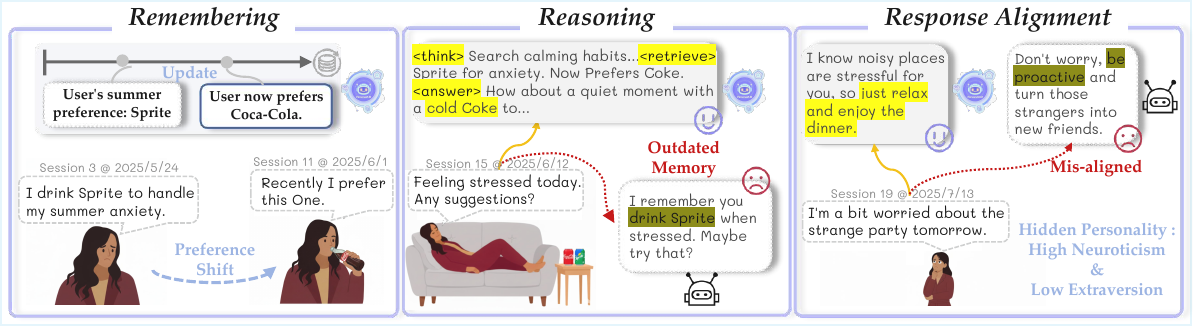} 
\caption{
Illustration of PersonaVLM's three core capabilities for long-term personalization.
PersonaVLM proactively remembers user preference shifts, performs multi-turn reasoning with retrieval, and generates responses aligned with the user's personality. In contrast, existing personalization strategies, such as input augmentation and output alignment, will result in poor recommendations based on outdated memories and replies that are misaligned with the user's personality.}
\label{img1}
\end{figure*}

Even advanced proprietary models exhibit limited capabilities in generating responses that cater to a user's unique preferences and characteristics~\cite{chen2024large,zhao2025llms,jiang2025know}.
This challenge stems from two primary factors: on the model side, they are predominantly optimized within fixed windows and a one-size-fits-all paradigm~\cite{li20251}; on the user side, an individual's preferences and personality are inherently diverse and dynamic, continuously evolving throughout ongoing interactions~\cite{jiang2025know}.
As illustrated in Fig.~\ref{img1}, a user initially expresses a preference for \textit{Sprite} but subsequently shifts to \textit{Coca-Cola} to mitigate anxiety in a multimodal interaction. When the user later expresses stress, a retrieval-augmented response fails to capture this shift, resulting in a misaligned recommendation.
Furthermore, a generic aligned response may feel overly extraverted, failing to accommodate the introverted and neurotic user whose personality traits are often revealed subtly across many unrelated dialogues.

The root of these failures is that current personalization strategies are designed for static interactions.
Specifically, input augmentation-based MLLMs like Yo’LLaVA~\cite{nguyen2024yo} and RAP~\cite{hao2025rap} 
specialize in recognizing user-specific concepts, but lack mechanisms to manage or update these memories, consequently failing to capture preference shifts from \textit{Sprite} to \textit{Coca-Cola}. Similarly, alignment techniques such as ALIGNXPERT~\cite{li20251} and Personality-Activation Search (PAS)~\cite{zhu2024personality} presuppose static user traits, preventing them from adapting to a user's introversion revealed contextually over time.
Therefore, we identify two foundational pillars for effective long-term personalization:
(\romannumeral1) \textit{\textbf{Personalized Memory Architecture}}. The ability to proactively construct and manage a dynamic, user-centric multimodal database.
(\romannumeral2) \textit{\textbf{Memory Utilization and Response Alignment}}.
The capacity to effectively utilize this database, employing reasoning and retrieval to generate responses that are deeply aligned with the user's unique and evolving characteristics.

Building on these pillars, we propose \textbf{PersonaVLM}, an innovative agent framework for long-term personalized interaction.
First, we design a memory architecture that integrates a user personality profile and four distinct memory types (\textit{core} for foundational attributes, \textit{semantic} for facts, \textit{procedural} for habits, and \textit{episodic} for events) to store and manage user information.
Second, building upon this architecture, a two-stage collaborative process transforms a general MLLM into a personalized assistant:
(1) Response stage: Given the user's multimodal input and context, PersonaVLM autonomously performs multi-step reasoning and memory retrieval to generate a response aligned with the user's personality.
(2) Update stage: The model infers and updates the user's latent traits, quantified as Big Five scores\footnote{
We represent user personality using the Big Five traits~\cite{roccas2002big}: Openness, Conscientiousness, Extraversion, Agreeableness, and Neuroticism (OCEAN), with each trait scored from 1 to 5.\label{fn:bigfive}}, through a momentum-based Personality Evolving Mechanism (PEM).
Concurrently, it proactively extracts and summarizes key knowledge from the dialogue, updating the four memory types for future use.
This integrated design endows PersonaVLM with the three key capabilities shown in Fig.~\ref{img1}.

Alongside the design of the framework, we address the scarcity of suitable training data by developing a synthesis pipeline to generate a large-scale personalized, multimodal interactive dataset, comprising over 30$k$ interactions across 500 unique personas.
This self-contained dataset enables effective training while ensuring PersonaVLM can operate locally, thereby eliminating data privacy concerns.
Furthermore, recognizing that existing benchmarks~\cite{liu2025survey} are often static and text-centric, we establish \textbf{Persona-MME}, a comprehensive benchmark designed to evaluate the long-term, multi-faceted, and multimodal personalization of MLLMs.
In summary, our contributions are fourfold:
\begin{itemize}
    \item We propose PersonaVLM, an innovative agent framework that achieves long-term personalization for MLLMs by integrating three core capabilities: proactive \textit{\textbf{R}emembering}, multi-step \textit{\textbf{R}easoning}, and \textit{\textbf{R}esponse Alignment}.

    \item We introduce a personalized memory architecture featuring two key components: the PEM for dynamic alignment and a multi-type memory database comprising core, procedural, semantic, and episodic memories.

    \item We establish Persona-MME, a comprehensive benchmark designed to evaluate the long-term and multi-faceted personalization capabilities of MLLMs, and use it to benchmark over 10 leading proprietary and open-source models.
    
    \item We conduct extensive experiments to validate the effectiveness of PersonaVLM. Under a 128$k$ context, PersonaVLM achieves improvements of 22.4\% on Persona-MME and 9.8\% on PERSONAMEM~\cite{jiang2025know}. Notably, it surpasses GPT-4o on these benchmarks and in open-ended evaluations.    
\end{itemize}

\section{Related Work}
\label{sec:rele}

\definecolor{figBlue}{HTML}{5071C7}
\definecolor{figPink}{HTML}{D49F9D}
\begin{figure*}[t]
\centering
\includegraphics[width=1.0\textwidth]{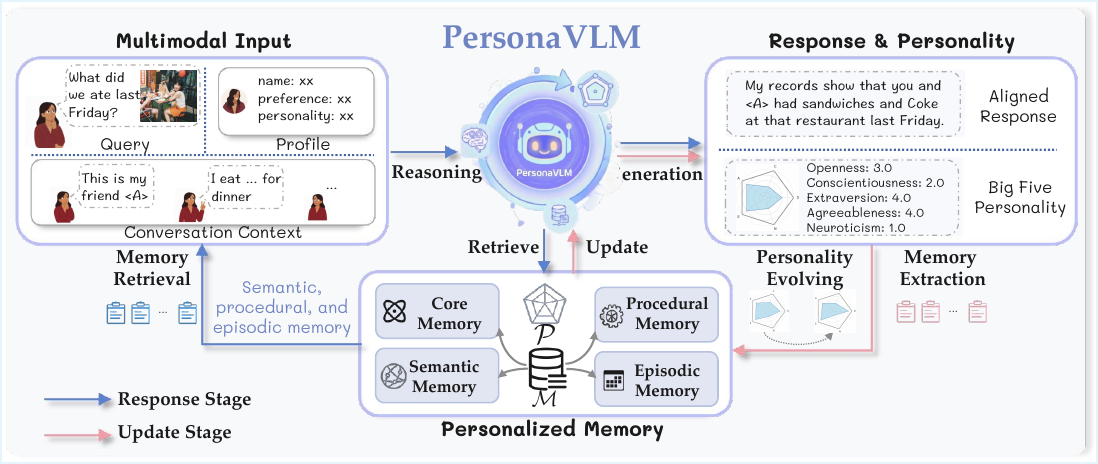} 
\caption{Overview of the PersonaVLM Framework. It leverages a personalized memory architecture and operates in two collaborative stages to achieve long-term personalization. In the Response Stage (\textcolor{figBlue}{blue arrows}), it processes multimodal input, retrieves from personalized memory, and generates a personality-aligned response. Subsequently, in the Update Stage (\textcolor{figPink}{pink arrows}), the framework analyzes the completed interaction to extract key memories and update the user's evolving personality profile$^{\ref{fn:bigfive}}$.\vspace{-2mm}}
\label{img2}
\end{figure*}

The recent surge in LLM development has catalyzed the emergence of powerful MLLMs like GPT-4o~\cite{hurst2024gpt}, LLaVA~\cite{liu2023visual}, and the Qwen series~\cite{bai2025qwen2,yang2025qwen3}, showcasing exceptional capabilities in various general-domain tasks~\cite{yin2024survey}.
However, to evolve into a true personal assistant, a model must transcend the ``one-size-fits-all'' paradigm and tailor responses to individual user knowledge and preferences~\cite{zhang2024personalization,liu2025survey}.
Existing efforts to address this challenge can be categorized into three primary streams: adaptation-based, augmentation-based, and alignment-based personalization.

\vspace{-2mm}
\noindent\paragraph{Adaptation-based Personalization.}
Adaptation-based methods operate at the model level, encoding user-specific knowledge directly into trainable parameters through fine-tuning.
Some works, for instance, employ parameter-efficient fine-tuning (PEFT) to adapt LLMs for individual users or groups~\cite{tan2024democratizing, zhuang2024hydra}.
This principle extends to the multimodal domain, where personalized MLLMs like MyVLM~\cite{alaluf2024myvlm} and Yo'LLaVA~\cite{nguyen2024yo} utilize learnable embeddings and soft prompts, respectively,  to represent user-specific visual concepts.
Such adaptation enables the model to transition from recognizing ``a generic dog'' to recognizing ``the user's pet dog.''
However, their reliance on fine-tuning for each new user concept renders these methods less scalable and unable to capture the evolution of user preferences.

\vspace{-2mm}
\noindent\paragraph{Augmentation-based Personalization.}
In contrast to model-level adaptation, augmentation-based approaches operate at the input level by equipping models with an external database to retain and retrieve user-specific memories~\cite{wang2023augmenting,wei2025ai}.
This paradigm is pivotal for transcending the limitations of fixed context windows in lifelong dialogues~\cite{chhikara2025mem0}.
Related approaches~\cite{hao2025rap,oh2025repic} extend personalization to the multimodal domain. They first employ open-vocabulary object detectors~\cite{liu2024grounding} to crop predefined visual concepts from images, which are then used for subsequent matching and retrieval.
A key advantage of these methods is their training-free nature\footnote{Following the specific terminology from~\cite{pi2024personalized}, this denotes that new user concepts can be accommodated at inference time without requiring continual fine-tuning.}, allowing them to accommodate new user concepts at inference time.
However, they are limited by a manually predefined database and lack mechanisms to proactively manage and update knowledge from dynamic interactions.
Moreover, while general-purpose memory architectures like A-Mem~\cite{xu2025mem} and Memory OS~\cite{li2025memos} employ more sophisticated agentic frameworks, their utility in our context is severely constrained. Their primary focus on text-only data limits their applicability to truly multimodal inputs, and their reliance on proprietary models creates barriers for open research and raises significant privacy concerns.

\vspace{-2mm}
\noindent\paragraph{Alignment-based Personalization.}
While standard LLM alignment, such as Reinforcement Learning from Human Feedback (RLHF)~\cite{ouyang2022training}, enforces a universal, ``one-size-fits-all'' behavioral standard, it inherently fails to accommodate diverse user preferences and communication styles.
As shown in Fig.~\ref{img1} (right), an overly enthusiastic response, while generally helpful, might be inappropriate for an introverted user experiencing anxiety.
Personalized alignment directly tackles this limitation by redefining the optimization objective from a universal standard to a user-specific one~\cite{liu2025survey}.
For example, Li et al.~\cite{li20251} incorporate user features into the input and use methods such as Direct Preference Optimization (DPO)~\cite{rafailov2023direct} to align model responses with predefined user values.
Another strategy, PAS~\cite{zhu2024personality}, trains user-specific ``probes'' to guide personalization at inference time. While this approach enables inference-time adaptation, it is fundamentally limited. Its reliance on per-user training poses significant scalability challenges; moreover, the static nature of these probes means the alignment can become outdated as the user's personality evolves over long-term interactions.

\vspace{2mm}
Departing from prior works that address siloed aspects of personalization for MLLMs, such as static memory or fixed alignment, we introduce PersonaVLM: a unified agent framework designed for dynamic, long-term interaction.

\section{Methods}
\label{sec:method}
\subsection{PersonaVLM Framework}
The overall architecture of the PersonaVLM agent is illustrated in Fig.~\ref{img2}. It is built upon a personalized memory architecture and operates through two collaborative stages of \textbf{Response} and \textbf{Update} to enable long-term personalization.

\vspace{-4mm}
\noindent\paragraph{Personalized Memory Architecture.}
This architecture is designed to construct and maintain a comprehensive, long-term user profile, storing two primary categories of information.
First, it maintains a {user personality profile} ($\mathcal{P}$), which provides a quantitative representation of the user's personality as a vector of scores for the Big Five dimensions\footnote{Representing user personality via the Big Five traits is a prevalent approach in LLM alignment~\cite{zhu2024personality}, rooted in psychological theories~\cite{john1999big,roccas2002big}.} (Openness, Conscientiousness, Extraversion, Agreeableness, and Neuroticism).
Second, it features a multi-type memory database ($\mathcal{M}$) that captures a wide range of user-related knowledge. This timeline-based, agentic system supports flexible CRUD (create, read, update, delete) operations and is structured into four distinct memory types:
\begin{itemize}
    \item \textbf{Core Memory:} Stores the user's fundamental attributes (e.g., \texttt{human} and \texttt{persona} blocks), inspired by MemGPT~\cite{packer2023memgpt}, and is dynamically updated to reflect their most current profile.
    \item \textbf{Semantic Memory:} Distills event-independent, abstract knowledge by extracting key entities, relationships, and multimodal concepts.
    \item \textbf{Episodic Memory:} Organizes raw dialogues into atomic, time-stamped events, each including a summary, dialogue turns, and keywords for efficient retrieval.
    \item \textbf{Procedural Memory:} Records user-centric plans, goals, and recurring behaviors or habits.
\end{itemize}
Regarding their storage and persistence, while episodic and semantic memories are stored chronologically, core and procedural memories, along with the personality profile, retain only their latest versions to ensure relevance.
Our design overcomes the limitations of existing systems, making our memory architecture: (a) Self-contained, avoiding proprietary model dependencies; (b) Explicitly personalized, prioritizing user-centric knowledge; and (c) Multimodal support, enabling a more holistic user understanding. 
For details on our memory architecture, refer to Appendix~\ref{sec:memo_appendix}.

\vspace{-4mm}
\noindent\paragraph{Response Stage.}
The objective of this stage is to generate an aligned response by performing multi-step reasoning and timeline-based retrieval. Formally, this process at turn $m$ can be formulated as:
\begin{equation}
\mathcal{R}_m = {R}(\mathcal{Q}_m, \mathcal{C}_{m},\mathcal{M}_{m-1}),
\label{eq1}
\end{equation}
where $\mathcal{R}_m$ is the personalized response. This response is conditioned on three inputs: the current user query $\mathcal{Q}_m = (T_m, I_m, t_m)$, consisting of a text instruction $T_m$, an optional image $I_m$, and a timestamp $t_m$; the dialogue context\footnote{We treat the recent conversation history (within a $t_{s}=60$ minute threshold) as short-term memory, and user inactivity beyond this threshold initiates a new session.} $\mathcal{C}_{m} = \left\{ (\mathcal{Q}_i, \mathcal{R}_i) \;\middle|\; 0 < i < m \text{ and } |t_i - t_m| \leq t_s \right\}$; and the state of the personalized memory database $\mathcal{M}_{m-1}$.
As depicted in the left panel of Fig.~\ref{img2}, the implementation of Eq.~(\ref{eq1}) is structured as a multi-step  interaction between the PersonaVLM agent and its memory system.
In the initial step, the model is prompted with the user's instruction, context, and a consolidated profile (comprising the user's core memory and personality).
The model then outputs a detailed reasoning process and an $\mathtt{action}$ result. If the model determines that the current information is insufficient, it outputs retrieval conditions within a predefined template, including the $\mathtt{time\ period}$ and $\mathtt{keywords}$ for searching.
The agent then executes the retrieval process by first isolating memories within the inferred $\mathtt{time\ period}$ and then performing a parallel search across semantic, episodic, and procedural memory types. The top-$k$ results from each type are collected and fed back to the model to initiate the next reasoning step. This iterative process continues for multiple rounds until the model outputs the final response $\mathcal{R}_m$.

Two key insights drive the design of this stage.
First, user queries are often highly context-dependent and contain anaphora (e.g., ``that thing we just talked about''), which renders direct semantic retrieval imprecise. In contrast, a multi-turn, agentic retrieval process typically yields more precise and efficient results~\cite{long2025seeing,jin2025search}.
Second, while some memory mechanisms~\cite{li2025memos,wang2025mirix} may leverage query rewriting~\cite{ma2023query} to improve retrieval accuracy, they overlook crucial temporal cues (e.g., ``this morning'').
Our design addresses these gaps by enabling the model to determine not just \textit{what} to retrieve, but also \textit{if} retrieval is necessary and from \textit{when}.

\begin{figure*}[!t]
\centering
\includegraphics[width=1.0\textwidth]{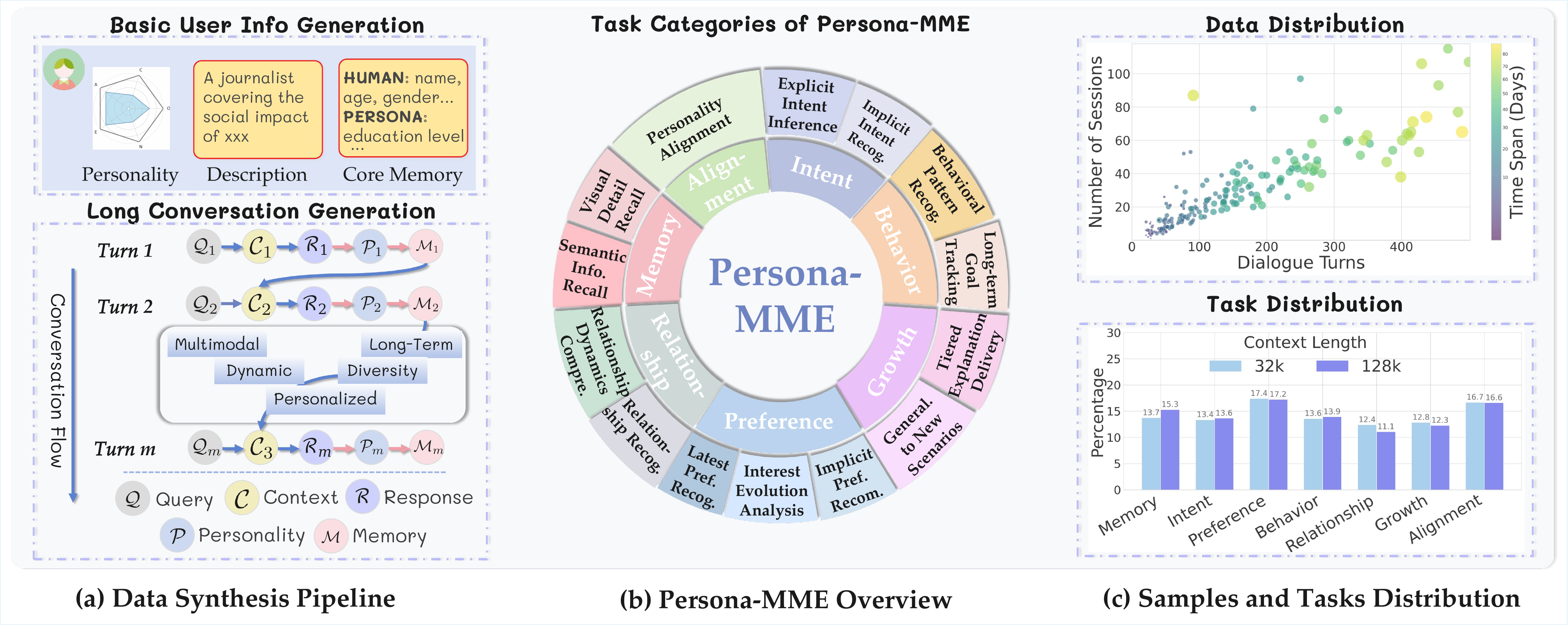} 
\caption{
Overview of our data synthesis pipeline and Persona-MME. (a) The pipeline first constructs rich user personas and then simulates long-term, dynamic conversations, generating both the dialogue and intermediate memories. (b) Persona-MME provides a comprehensive evaluation of personalization by assessing 14 fine-grained capabilities. (c) Statistics for Persona-MME, which includes two context length configurations (32$k$ and 128$k$) and contains over 2,000 \textit{in-situ}$^{\ref{foot:insitu}}$ cases.\vspace{-5mm}}
\label{img3}
\end{figure*}

\vspace{-4mm}
\noindent\paragraph{Update Stage.}
This stage, which executes automatically during idle periods after a response is generated, primarily involves two parts: evolving the user's personality profile and proactively updating the memories.
This process at turn $m$ can be represented as:
\begin{equation}
(\mathcal{P}_m, \mathcal{M}_m) = {U}(\mathcal{Q}_m,\mathcal{R}_m,\mathcal{M}_{m-1}).
\label{eq2}
\end{equation}
Specifically, the user's personality profile, $\mathcal{P}_m$, is updated via our proposed \textbf{Personality Evolving Mechanism (PEM)}. 
The PEM maintains a long-term personality profile as a vector $\mathbf{p} \in \mathbb{R}^5$, corresponding to the Big Five dimensions~\cite{zhu2024personality}.
At each turn $m$, the PEM first infers a temporary set of personality scores from the user's latest query, $\mathcal{Q}_m$. 
These scores are normalized to form a turn-specific personality vector, $\mathbf{p}'_m$. 
Subsequently, the long-term profile vector is updated using an exponential moving average (EMA):
$\mathbf{p}_{m} \leftarrow \lambda \cdot \mathbf{p}_{m-1} + (1 - \lambda) \cdot \mathbf{p}'_m$,
where $\lambda \in [0, 1]$ is a dynamic smoothing factor. 
To ensure high adaptability in early conversations while promoting stability over time, we employ a cosine decay schedule for $\lambda$.
It starts with a low value (allowing rapid adaptation to initial user interactions) and gradually increases, making the profile more stable and less susceptible to minor fluctuations.
Finally, the updated numerical vector $\mathbf{p}_{m}$ is converted back into a descriptive textual summary, $\mathcal{P}_m$, for use in the Response Stage.

Second, we selectively extract and update the four memory types, each with tailored logic. 
Semantic memory is updated after each turn, where key information such as user preferences, multimodal concepts, and explicit memorization requests is extracted and stored with timestamps and keywords. 
In contrast, core and procedural memory are updated at the end of each session; the agent analyzes the entire session's dialogue to perform automated CRUD operations and keep these memories current.
Finally, episodic memory is constructed by segmenting dialogues into distinct topics, with each entry containing a summary, relevant keywords, and the specific dialogue turns involved.
See Appendix~\ref{sec:pr3detail:pipeline} for the complete implementation pipeline.

\subsection{Training of PersonaVLM}
We adopt Qwen2.5-VL-7B~\cite{bai2025qwen2} as the backbone model for PersonaVLM and train it using a two-stage process. 

\vspace{-4mm}
\noindent\paragraph{Stage 1: Supervised Fine-Tuning (SFT).}
We perform SFT on a curated synthetic dataset of 78$k$ samples to equip the model with foundational memory management and multi-turn reasoning skills. The training data is synthesized via a pipeline introduced in the next section and comprises two primary types: (a) examples for memory mechanisms, including personality inference and the four types of memory CRUD operations; and 
(b) QA pairs containing complete, multi-step reasoning trajectories constructed offline. After SFT, the model is capable of generating well-formed reasoning and retrieval actions, providing a strong cold-start initialization for the subsequent stage.

\vspace{-4mm}
\noindent\paragraph{Stage 2: Reinforcement Learning (RL).}
This stage aims to further enhance the model's multi-turn reasoning capability. We employ Group Relative Policy Optimization (GRPO)~\cite{guo2025deepseek}, an improved PPO algorithm, to train the policy model $\pi_{\theta}$. During generation, we enforce a strictly structured output format: the model must first output its reasoning process within \texttt{<think></think>} tags, followed by either retrieval conditions in \texttt{<retrieve></retrieve>} tags or the final response in \texttt{<answer></answer>} tags. For each training sample $\{\mathcal{Q}, \widehat{\mathcal{R}}\}$, where $\mathcal{Q}$ is the user input and $\widehat{\mathcal{R}}$ is the preferred response, a group of multi-turn trajectories $\{\tau_1, \dots, \tau_G\}$ is sampled from the policy model. The reward for the $i$-th trajectory $\tau_i$ is calculated as:
\begin{equation}
    r_i = f_{\text{acc}}(\widehat{\mathcal{R}}, \mathcal{R}_{\tau_i}) \cdot f_{\text{cons}}(\mathcal{Q}, \mathcal{R}_{\tau_i}) + 0.5 \cdot f_{\text{format}}(\mathcal{R}_{\tau_i}),
    \label{eq:reward}
\end{equation}
where $f_{\text{acc}}$, $f_{\text{cons}}$, and $f_{\text{format}}$ are reward functions for accuracy, logical consistency between reasoning and the final answer, and format adherence, respectively.
We use Qwen3-30B-A3B~\cite{yang2025qwen3} as an \textit{LLM-as-a-Judge} to compute $f_{\text{acc}}$ and $f_{\text{cons}}$ via zero-shot prompting.
Following~\cite{guo2025deepseek}, the advantage for each trajectory is computed by standardizing its reward within the sampled group.
During training, we cap the maximum number of retrieval attempts at three per trajectory, and the loss is computed exclusively on the generated tokens.
Further details on the training data and implementation are provided in Appendix~\ref{sec:pr3detail:training}.

\section{Dataset and Persona-MME Construction}
\label{sec:data}

To enable both the implementation and evaluation of long-term dynamic personalization, we make two key contributions. First, to address the scarcity of high-quality training data, we construct a large-scale multimodal interaction dataset via a dedicated synthesis pipeline.
Second, we establish Persona-MME, a comprehensive benchmark for evaluating personalization in multimodal settings.
This dual effort is necessitated by existing datasets~\cite{nguyen2024yo,li20251}, which are typically static, single-turn, or lack multimodal support.

\vspace{-2mm}
\noindent\paragraph{Dataset Synthesis Pipeline.}
As illustrated in Fig.~\ref{img3} (a), we design a synthesis pipeline to generate training data at \textit{scale}.
The process commences by sampling base personas from PersonaHub~\cite{ge2024scaling}, which are then enriched with randomly assigned personality traits. This enrichment step generates a detailed role description and an initial user profile, forming the initial Core Memory.
We employ Seed1.6-thinking\footnote{Seed1.6-thinking is a commercial model with performance comparable to GPT-4o, selected for its balance of capability and cost-effectiveness.} to generate conversations guided by a structured flow.
This process is governed by several key principles:
(1) \textbf{Long-term Dynamics}: Dialogues extend over hundreds of turns to simulate interactions spanning weeks or months. To capture this longitudinal evolution, we probabilistically induce dynamic shifts in user preferences, topics, and personality traits.
(2) \textbf{Multimodality and Scenario Diversity}: Over 15\% of dialogues incorporate multimodal elements. The interactions span a wide range of real-world scenarios, from professional tasks to casual conversations.
(3) \textbf{Structured Supervision}: The generation process is guided to produce not only the conversational dialogue but also the intermediate reasoning, retrieval, and memorization steps. This explicit structure provides rich supervisory signals for training the PersonaVLM framework.
Further details on the data distribution and validation process are provided in Appendix~\ref{sec:data_curation}.

\vspace{-2mm}
\noindent\paragraph{Persona-MME: Evaluating Long-Term Personalization of MLLMs.}
Existing benchmarks focus on siloed aspects of personalization. For instance, PERSONAMEM~\cite{jiang2025know} evaluates a model's ability to track a user's evolving profile, ALIGNX-test~\cite{li20251} is centered on static alignment, and others like Yo'LLaVA~\cite{nguyen2024yo,hao2025rap} assess user-specific concept understanding. However, none provide a holistic evaluation across the critical dimensions of dynamic personalization.

To fill this void, we introduce Persona-MME, a comprehensive benchmark comprising over 2,000 \textit{in-situ}\footnote{Queries are posed from the user's first-person perspective at a specific point in the conversational history, simulating a realistic interaction~\cite{jiang2025know}\label{foot:insitu}.} cases derived from 200 diverse personas.
As depicted in Fig.~\ref{img3} (b), Persona-MME is structured around seven core dimensions: \textbf{Memory, Intent, Preference, Behavior, Relationship, Growth, and Alignment}. Together, these dimensions encompass 14 fine-grained tasks, which are detailed in Table~\ref{tab:task_definitions} in the Appendix.
To accommodate different context lengths, we provide two evaluation configurations: a 32$k$-context version for dialogues under 100 turns and a 128$k$-context version for longer interactions, each containing cases from 100 distinct personas.
Each test case comprises (1) a multiple-choice question assessing the model's personalized memory and understanding, and (2) an optional personality test evaluating its alignment.
This multi-faceted structure enables Persona-MME to evaluate an MLLM's long-term personalization capabilities across diverse personas.
Further details and statistics are provided in Appendix~\ref{sec:bench_detail}.

\begin{figure}[!t]
\centering
\includegraphics[width=.5\textwidth]{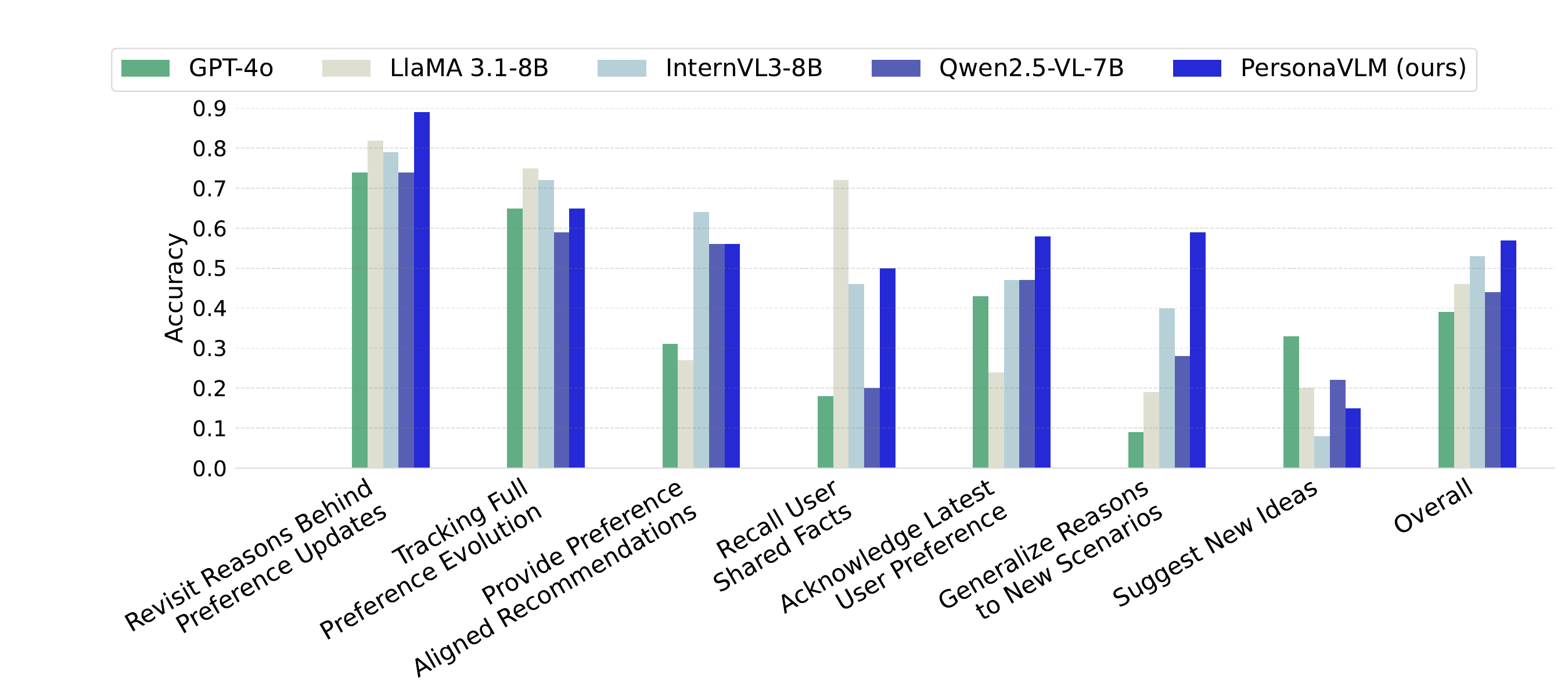}
\caption{Quantitative evaluation across seven tasks on the PERSONAMEM (32$k$) benchmark.\vspace{-5mm}}
\label{img4}
\end{figure}

\begin{table*}[!htbp]
\centering
\caption{
Evaluation on the Persona-MME and PERSONAMEM benchmarks, tested at context lengths of 32$k$ and 128$k$.
We report accuracy (\%) for Persona-MME (overall and across six aspects) and PERSONAMEM. The comparison includes two settings: full-context (``Full'') and retrieval-augmented generation (``RAG'').
Best results are shown in \textbf{bold}.
The GPT-4o results on PERSONAMEM are from~\cite{jiang2025know}.
    }
\label{tab:tab1_eval_pr3}
\resizebox{0.9\textwidth}{!}{
\begin{tabular}{c|l|cccccc|c|c}
\toprule
\multirow{2}{*}{\textbf{Context}}    & \multicolumn{1}{c}{\multirow{2}{*}{\textbf{Model}}} & \multicolumn{7}{c}{\textbf{Persona-MME}}    & \multicolumn{1}{c}{\multirow{2}{*}{\textbf{PERSONAMEM}}} \\
\cmidrule(l){3-9}
& \multicolumn{1}{c}{}   & Memory & Intent & Preference & Behavior & Relationship & Growth & Overall &            \\
\midrule
    & GPT-4o     &   \textbf{86.99} & \textbf{83.87} & \textbf{63.12} & 57.14 & 71.30 & 73.87 & \textbf{72.35}    &     39.20        \\
\cmidrule(l){2-10}
\rowcolor{gray!15}
 \cellcolor[HTML]{FFFFFF}\multirow{2}{*}{32$k$-Full}  & Qwen2.5-VL-7B &   66.13 & 66.85 & 59.75 & 59.24 & 68.45 & 70.69 & 64.84    &    43.63        \\
  & InternVL3-8B   &   56.45 & 76.24 & 57.20 & 54.35 & 69.05 & 74.14 & 64.04     &      52.97      \\
 & InternVL3-38B    &    66.67 & 85.64 & 66.53 & 59.78 & 72.02 & 77.59 & 71.04    &     \textbf{57.93}       \\
  & OneVision-1.5-8B       &    74.19 & 74.59 & 60.59 & 53.26 & 72.62 & 74.14 & 67.76    &     52.80       \\
\cmidrule(l){1-10}
 & Qwen2.5-VL-7B & 65.05 & 68.51 & 50.42 & 57.61 & 60.71 & 68.39 & 61.20 &    45.67        \\
\rowcolor{blue!10}
\cellcolor[HTML]{FFFFFF}\multirow{1}{*}{32$k$-RAG}& PersonaVLM$_{\text{SFT}}$      &  67.20 & 70.17 & 49.58 & 57.07 & 70.24 & 80.46 & 64.84{\color{gray}$_{+3.64}$}     &     52.12{\color{gray}$_{+6.45}$}       \\
\rowcolor{blue!15}
\cellcolor[HTML]{FFFFFF}  & PersonaVLM$_{\text{RL}}$   &    69.89 & 76.80 & 58.05 & \textbf{69.02} &\textbf{ 73.21} & \textbf{86.78} & 71.48{\color{gray}$_{+10.28}$}     &      56.53{\color{gray}$_{+10.86}$}      \\
\midrule 
\midrule
   & GPT-4o                  &   \textbf{84.44} & 75.63 & 59.12 & 55.65 & 65.98 & 76.64 & 69.23    &    45.32        \\
\cmidrule(l){2-10}
\cmidrule(l){2-10}
\rowcolor{gray!15}
 \cellcolor[HTML]{FFFFFF}\multirow{2}{*}{128$k$-Full}    & Qwen2.5-VL-7B     &    50.60 & 54.73 & 52.41 & 54.30 & 55.83 & 60.90 & 54.48    &    3.08       \\
  & InternVL3-8B      &   57.23 & 68.92 & 53.48 & 54.97 & 69.17 & 76.69 & 62.43    &      36.62      \\
& InternVL3-38B    &    67.47 & 71.62 & \textbf{64.71} & 58.94 & 65.00 & 76.69 & 67.18    &         46.56   \\
  & OneVision-1.5-8B    &    52.44 & 54.79 & 58.15 & 45.33 & 65.25 & 67.18 & 56.66    &     14.28       \\
\cmidrule(l){1-10}
\cmidrule(l){2-10}
 & Qwen2.5-VL-7B &  56.63 & 63.51 & 50.27 & 55.63 & 61.67 & 70.68 & 59.01     &    37.88     \\
\rowcolor{blue!10}
\cellcolor[HTML]{FFFFFF}\multirow{1}{*}{128$k$-RAG} & PersonaVLM$_{\text{SFT}}$      &     67.47 & 75.68 & 59.36 & 51.66 & 71.67 & 81.95 & 67.18{\color{gray}$_{+8.17}$}     &      43.60{\color{gray}$_{+5.72}$}       \\
\rowcolor{blue!15}
 \cellcolor[HTML]{FFFFFF}  & PersonaVLM$_{\text{RL}}$   &  69.28 & \textbf{77.70} & 61.50 & \textbf{60.26} & \textbf{75.00} & \textbf{87.97} & \textbf{71.05}{\color{gray}$_{+12.04}$}  &    \textbf{ 47.28}{\color{gray}$_{+9.4}$}       \\
\bottomrule
\end{tabular}}
\vspace{-2mm}
\end{table*}

\section{Experiments}
\label{sec:exp}
In this section, we present a series of quantitative and qualitative experiments designed to validate our PersonaVLM framework. The evaluation in the main paper is structured to answer the following research questions (RQs):

\begin{itemize}
    \item \textbf{RQ1:} How effectively does PersonaVLM perform in personalized user understanding and memory recall?

    \item \textbf{RQ2:} Can PersonaVLM achieve effective alignment by capturing a user's evolving personality traits over time?

    \item \textbf{RQ3:} How well does PersonaVLM perform in personalized open-ended generation?
\end{itemize}
For comprehensive evaluations of Persona-MME, ablation studies about memory components, and further discussions, please refer to Appendices~\ref{sec:bench_detail},~\ref{sec:exp_detail}, and~\ref{sec:discussion}, respectively.

\subsection{Personalized Understanding Evaluation}
To evaluate personalized understanding (RQ1), we conduct experiments on two benchmarks: our Persona-MME and PERSONAMEM~\cite{jiang2025know}. The latter includes seven task types specifically designed to assess a model's ability to track dynamic user preferences over the long term. We evaluate all models under two long-context settings (32$k$ and 128$k$ tokens), with detailed results reported in Table~\ref{tab:tab1_eval_pr3} and Fig.~\ref{img4}. For comparison, we benchmark against several powerful models, including the proprietary GPT-4o~\cite{hurst2024gpt} and strong open-source models such as Qwen2.5-VL-7B~\cite{bai2025qwen2}, LLaVA-OneVision-1.5-8B~\cite{an2025llava}, and InternVL3-8B/38B~\cite{zhu2025internvl3}.
See Appendix Fig.~\ref{fig:model_performance_comp} for more comparisons with leading models.

Compared to strong open-source models of a similar size, such as InternVL3-8B and LLaVA-OneVision-1.5-8B (provided with full context), PersonaVLM shows improvements of 8.62\% and 14.39\% on Persona-MME in the 128$k$ setting, respectively. 
While the personalization capabilities of these open-source models appear to improve with scale, PersonaVLM still outperforms the much larger InternVL3-38B by 3.87\% on Persona-MME (128$k$).
We also evaluate Qwen2.5-VL-7B augmented with a straightforward RAG setup, which retrieves the top five most relevant messages following the approach of~\cite{jiang2025know}.
Interestingly, the results show that RAG can be detrimental in short-context scenarios—degrading performance on preference understanding tasks by as much as 9.33\%—while providing a substantial boost of 4.53\% in long-context settings. Additionally, as shown in Table~\ref{tab:tab1_eval_pr3}, the two-stage training process demonstrates clear effectiveness, yielding an average improvement of 5.35\% on Persona-MME.

\begin{figure}[!t]
\centering
\includegraphics[width=.45\textwidth]{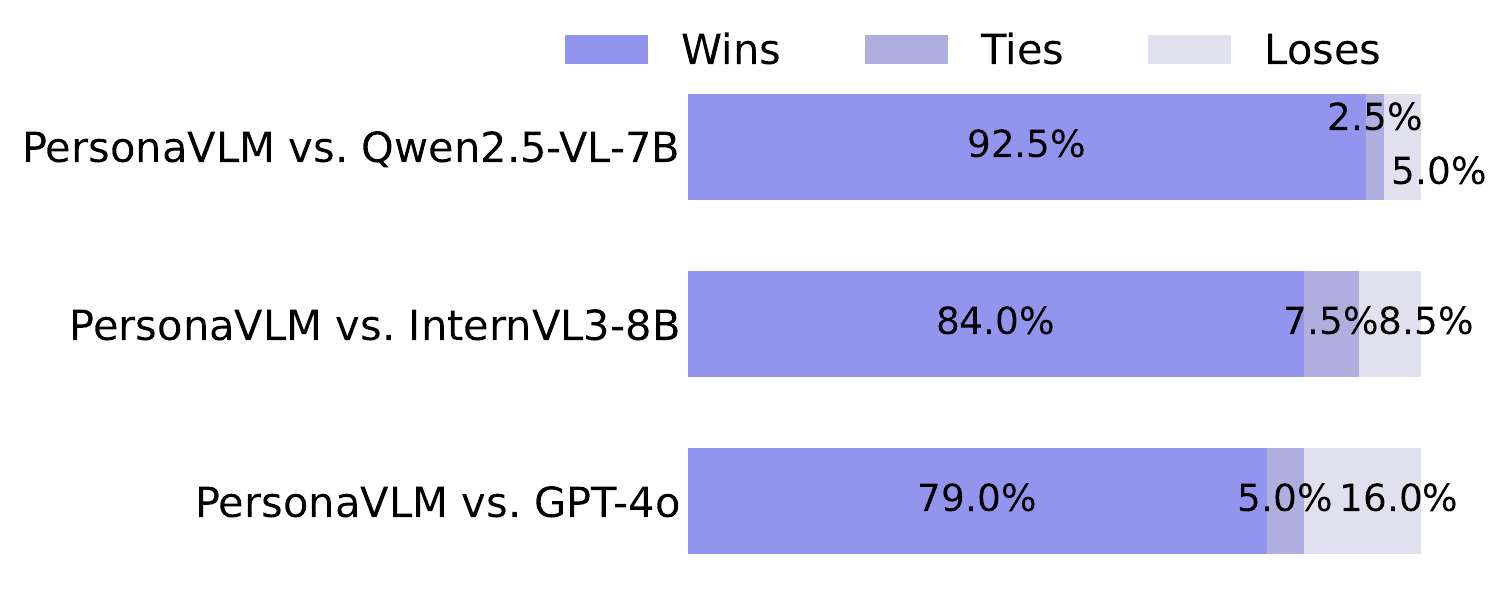} 
\caption{
Qualitative comparison on open-ended generation, evaluated by Gemini-2.5-Pro. The evaluation assesses both the factual accuracy and the personality alignment of the responses.\vspace{-5mm}}
\label{img6}
\end{figure}

When benchmarked against the proprietary GPT-4o, our method achieves competitive results on Persona-MME and demonstrates notable improvements of 17.3\% and 2.0\% on the 32$k$ and 128$k$ configurations of PERSONAMEM, respectively.
Furthermore, while PersonaVLM's performance in memory recall lags behind that of GPT-4o with full context—a finding consistent with~\cite{kang2025memoryosaiagent}—it demonstrates a significant advantage elsewhere. 
In particular, in Growth Modeling and Behavioral Awareness, PersonaVLM outperforms GPT-4o by over 10\%.

\begin{table}[!tbp]
    \centering
    \caption{Evaluation of personalized alignment on the Persona-MME and P-SOUPS benchmarks.}
    \label{tab:alignment}
\resizebox{0.5\textwidth}{!}{
    \begin{tabular}{l|cc|cccc}
    \toprule
        \multirow{2}{*}{\textbf{Model}} & \multicolumn{2}{c}{\textbf{Persona-MME}} & \multicolumn{4}{c}{\textbf{P-SOUPS}}\\
        \cmidrule(lr){2-3} \cmidrule(lr){4-7}
        &{32$k$} & {128$k$} & ${\text{Expertise}}$ & ${\text{Informativeness}}$ & ${\text{Style}}$ & ${\text{Overall}}$ \\
\midrule
Qwen2.5-VL-7B & 69.91 & 52.27 & 39.00 & 49.34 & 23.00 & 37.11 \\
InternVL3-8B & 55.75 & 55.56 & 47.16 & 51.81 &26.16 & 41.71 \\
InternVL3-38B & 64.60 & 63.01  &  \textbf{{52.80}} & 53.30 & 32.83 & 46.32 \\
Qwen3-30B-A3B & 80.09 & 83.06  & 51.67 & \textbf{54.60}  & 35.16 & 47.14 \\
\midrule
\multicolumn{7}{l}{\ \textit{baseline with different strategies} } \\
Self-Critic &59.73&57.66&39.67&49.33&23.67 & 37.50 \\
Few-Shot &-&-&42.16&48.66& 28.33& 39.67\\
\midrule
\rowcolor{blue!15}
 \cellcolor[HTML]{FFFFFF}
PersonaVLM (ours) & \textbf{89.16} & \textbf{ 92.22 }& 51.16 & {53.65} &\textbf{44.00}  & \textbf{49.60} \\        
\bottomrule
\end{tabular}}
\vspace{-5mm}
\end{table}

\begin{figure*}[!th]
\centering
\includegraphics[width=.9\textwidth]{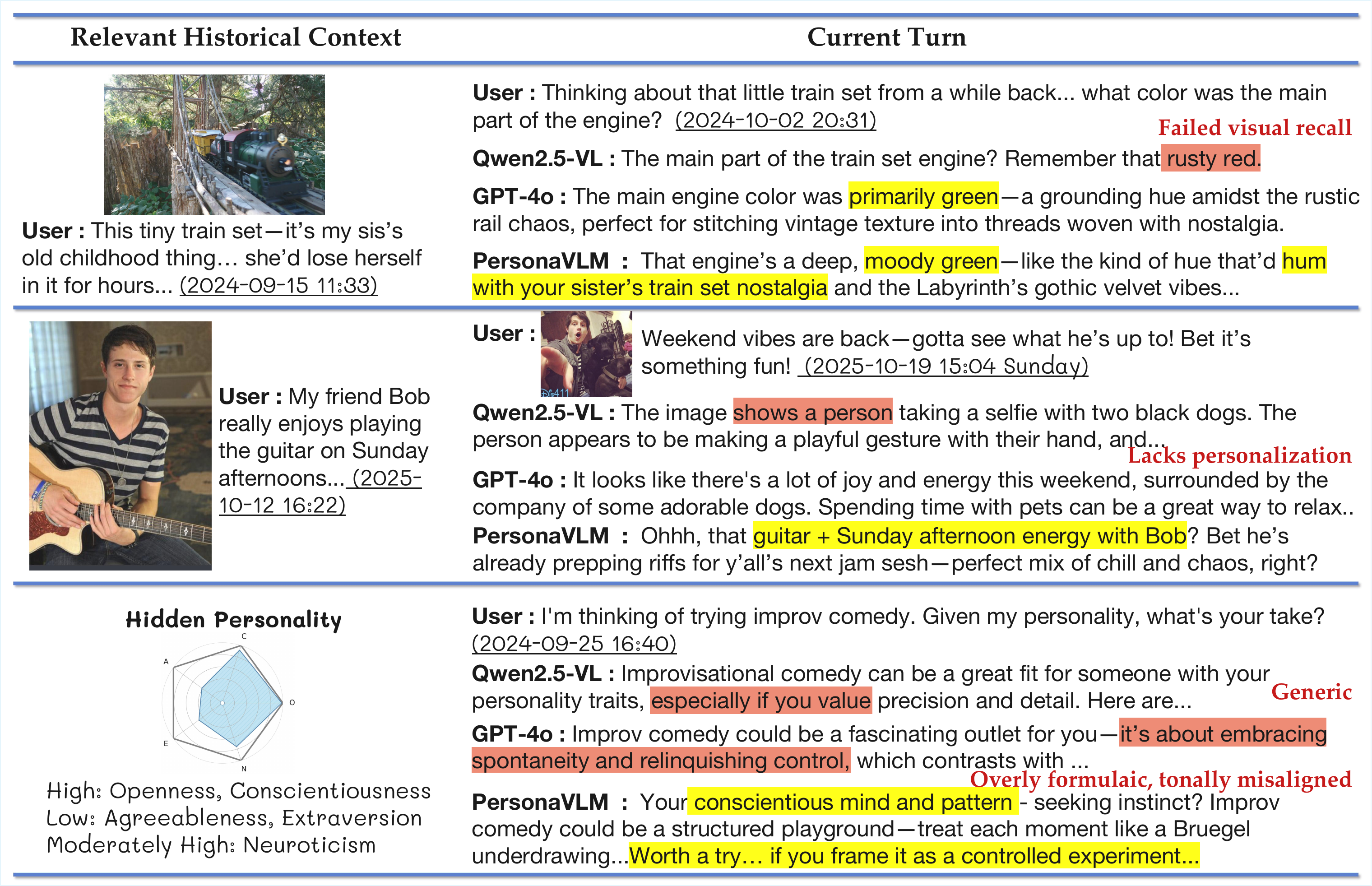} 
\caption{Qualitative comparison on open-ended generation tasks. Case studies demonstrate PersonaVLM's superior capabilities in memory recall, context integration, and personality alignment compared to the baseline and GPT-4o.\vspace{-4mm}}
\label{img5}
\end{figure*}

\subsection{Personalized Alignment Evaluation}
For RQ2, we conduct experiments on two benchmarks: the Alignment sub-task within Persona-MME and the P-SOUPS~\cite{jang2023personalized}, which comprise 812 and 1,800 test cases, respectively. The former assesses a model's ability to determine if a response aligns with a user's personality inferred from the conversational context. The latter evaluates personality alignment with a given user profile across three dimensions: Expertise, Informativeness, and Style.

We quantitatively compare PersonaVLM against several powerful open-source models, including InternVL3-8B/38B and Qwen3-30B-A3B~\cite{yang2025qwen3}, with the latter being noted for its strong language capabilities. We also evaluate the baseline model augmented with different strategies, such as Self-Critic and few-shot prompting~\cite{zhao2025llms}.
As shown in Table~\ref{tab:alignment}, PersonaVLM consistently outperforms existing models on both benchmarks. Notably, it leads the next-best model by 9.16\% on Persona-MME and 2.46\% on P-SOUPS, representing a $>$12\% gain over the baseline. Interestingly, language-centric models (e.g., Qwen3-30B-A3B) exhibit stronger alignment than multimodal counterparts like InternVL3-38B, with a 20\% margin on Persona-MME (128$k$).
These outcomes underscore PersonaVLM's capacity for robust personality alignment.

\subsection{Qualitative Evaluation}
To address RQ3 on open-ended generation, we conduct an automated evaluation using 200 questions randomly sampled from Persona-MME. We benchmark PersonaVLM against InternVL3-8B, Qwen2.5-VL-7B, and GPT-4o, employing Gemini-2.5-Pro~\cite{comanici2025gemini} as an automated judge.
Responses are assessed on two criteria: Accuracy and Personality Alignment, with PersonaVLM's performance in pairwise comparisons classified as a ``win,'' ``tie,'' or ``loss.''
The evaluation prompt is provided in Fig.~\ref{fig:prompt_evaluation}.
As illustrated in Fig.~\ref{img6}, PersonaVLM achieves a substantially higher win rate than its peers. Particularly striking is its head-to-head performance against GPT-4o, where PersonaVLM secures a 79\% win rate versus a 16\% loss rate. This is further corroborated by qualitative case studies in Fig.~\ref{img5}, which showcase PersonaVLM's ability to perform accurate visual recall, integrate contextual memory, and maintain long-term personality alignment. In contrast, other models exhibit critical failures, such as memory hallucinations or tonally misaligned responses that ignore user-specific memories. These findings validate the generative capabilities of PersonaVLM for long-term personalization.

\section{CONCLUSION}
This paper introduces PersonaVLM, a novel agent framework that enables long-term, dynamic personalization for MLLMs by integrating three core capabilities: Remembering, Reasoning, and Response Alignment.  
To support rigorous evaluation, we further propose Persona-MME, a comprehensive benchmark for personalized multimodal understanding.  
Experiments show that PersonaVLM significantly enhances a model's personalization capabilities and consistently outperforms strong counterparts, including both proprietary GPT-4o and leading open-source alternatives.
Our work provides a new paradigm for developing truly user-centric AI assistants, and future work will extend these capabilities toward a fully immersive multimodal experience.

{
    \small
    \bibliographystyle{ieeenat_fullname}
    \bibliography{main}
}
\clearpage
\setcounter{page}{1}
\setcounter{section}{0}
\maketitlesupplementary
\renewcommand{\thesection}{\Alph{section}}

\noindent This supplementary material provides comprehensive details to complement the main paper, organized as follows:

\begin{itemize}

    \item \textbf{Appendix~\ref{sec:memo_appendix}} elaborates on our proposed memory architecture, detailing each memory component—including its storage, retrieval, and update processes.
    
    \item \textbf{Appendix~\ref{sec:pr3detail}} outlines the training and implementation details of PersonaVLM framework.
    
    \item \textbf{Appendix~\ref{sec:data_curation}} presents a detailed analysis of our synthesized dataset, covering its distribution and the validation process.
    
    \item \textbf{Appendix~\ref{sec:bench_detail}} offers a comprehensive breakdown of Persona-MME, including its task taxonomy, detailed statistical analysis, and full evaluation results.

    \item \textbf{Appendix~\ref{sec:exp_detail}} presents additional experimental details, including ablation studies and the full set of prompts used in our framework.

    \item \textbf{Appendix~\ref{sec:discussion}} offers further efficiency analysis and limitations of PersonaVLM.
\end{itemize}

\section{Details of the PersonaVLM Memory Architecture}
\phantomsection
\label{sec:memo_appendix}
As introduced in Section~\ref{sec:method}, the PersonaVLM memory architecture comprises two components: a User Personality Profile ($\mathcal{P}$) and a Multi-Type Memory Database ($\mathcal{M}$). This section provides a detailed exposition of how these memories are stored, updated, and retrieved.

\subsection{Memory Storage}

\vspace{-2mm}
\noindent\paragraph{User Personality Profile ($\mathcal{P}$).}
We quantitatively represent the user's personality as a five-dimensional vector, $\mathbf{p} \in \mathbb{R}^5$, where each element corresponds to a Big Five trait and is a floating-point value between 1 and 5. This profile is dynamically updated after each interaction turn $m$.
Specifically, at the end of a turn, the model infers a personality vector, $\mathbf{p}'_m \in \mathbb{R}^5$, where each component is an integer score from 1 to 5 based on the user's current input and context. The persistent personality profile $\mathbf{p}$ is then updated using an Exponential Moving Average (EMA):
$
\mathbf{p} \leftarrow  \lambda_m\mathbf{p} + (1 - \lambda_m) \mathbf{p}'_m
$
where the smoothing factor $\lambda_m$ is dynamically adjusted to be more sensitive in early interactions and stabilize over time:
$
\lambda_m = 0.7 - 0.2 \cdot \cos\left(\frac{\min(m, 50)}{50} \pi\right).
$
To ensure stability, this update is applied selectively.
The process is skipped if the inferred personality vector $\mathbf{p}'_m$ consists solely of the neutral score (3), a condition that typically arises in non-personalized or neutral contexts.
During the response generation stage, the personality profile $\mathcal{P}$ is provided to the model via structured prompting.

\vspace{-2mm}
\noindent\paragraph{Core Memory.}
Core memory stores the user's foundational and high-priority attributes and is included in every interaction turn. It is divided into two sub-components~\cite{packer2023memgpt}:
\begin{itemize}
\item \textbf{Human:} Factual user attributes, such as age, gender, preferences, and interests, with the user's name as a mandatory field. This information provides PersonaVLM with a foundational understanding of the user's background.
\item \textbf{Persona:} The user's identity, roles (e.g., ``a meticulous researcher''), and explicit requirements for the model's interaction style, tone, and behavior.
\end{itemize}

\vspace{-2mm}
\noindent\paragraph{Semantic Memory.} Semantic memory~\cite{wang2025mirix} archives timeless, multimodal knowledge that is either explicitly provided by the user or autonomously inferred by the model. This knowledge is categorized as follows:
\begin{itemize}
\item \textbf{Explicit Directives:} Direct commands from the user to remember specific information, which can be textual or visual. For example, a user might provide an image and say, ``Remember the boy in this picture.''
\item \textbf{Core Facts:} Stable, factual information about the user disclosed during conversation, such as their profession, significant life events, or specific requirements for the agent's behavior.
\item \textbf{Preferences \& Habits:} User preferences for entities, visual styles, or activities, which can be either explicitly stated or implicitly revealed through behavior patterns.
\item \textbf{Visual Concepts:} User-specific visual concepts that arise in multimodal dialogues, such as friends, pets, or personal items. These are stored as a key-value pair linking a textual description to an image crop, formatted as ``simple description <image>''.
\end{itemize}
Beyond these predefined categories, the agent autonomously determines at the end of each turn whether new semantic knowledge warrants storage. If so, it generates a structured output containing the reasoning process, memory content, and a set of keywords for future retrieval.

\vspace{-2mm}
\noindent\paragraph{Episodic Memory.}
Episodic memory archives both summaries and raw data from past conversations. For each multi-turn dialogue session, the model segments the conversation by topic. Each resulting topic-based episode contains three key elements: (a) a concise summary, (b) a set of keywords, and (c) the indices of the dialogue turns constituting that episode. To ensure no details are lost, the original dialogue data is never deleted; the episodic memory serves as a structured layer for organizing and retrieving this raw data.

\vspace{-2mm}
\noindent\paragraph{Procedural Memory.}
Procedural memory tracks user goals and identifies recurring behaviors or habits by storing procedural events from conversations. It primarily stores two types of information:
\begin{itemize}
\item \textbf{Long-term Goals:} Ongoing projects, plans, or objectives that the user is working towards.
\item \textbf{Habits \& Routines:} Repetitive behaviors or workflows that are automatically identified from user interactions.
\end{itemize}
Similar to Core Memory, this information is stored as key-value pairs, and only the latest version is retained.

\begin{figure}[!t]
    \centering
    \begin{subfigure}{0.5\textwidth}
        \includegraphics[width=\textwidth]{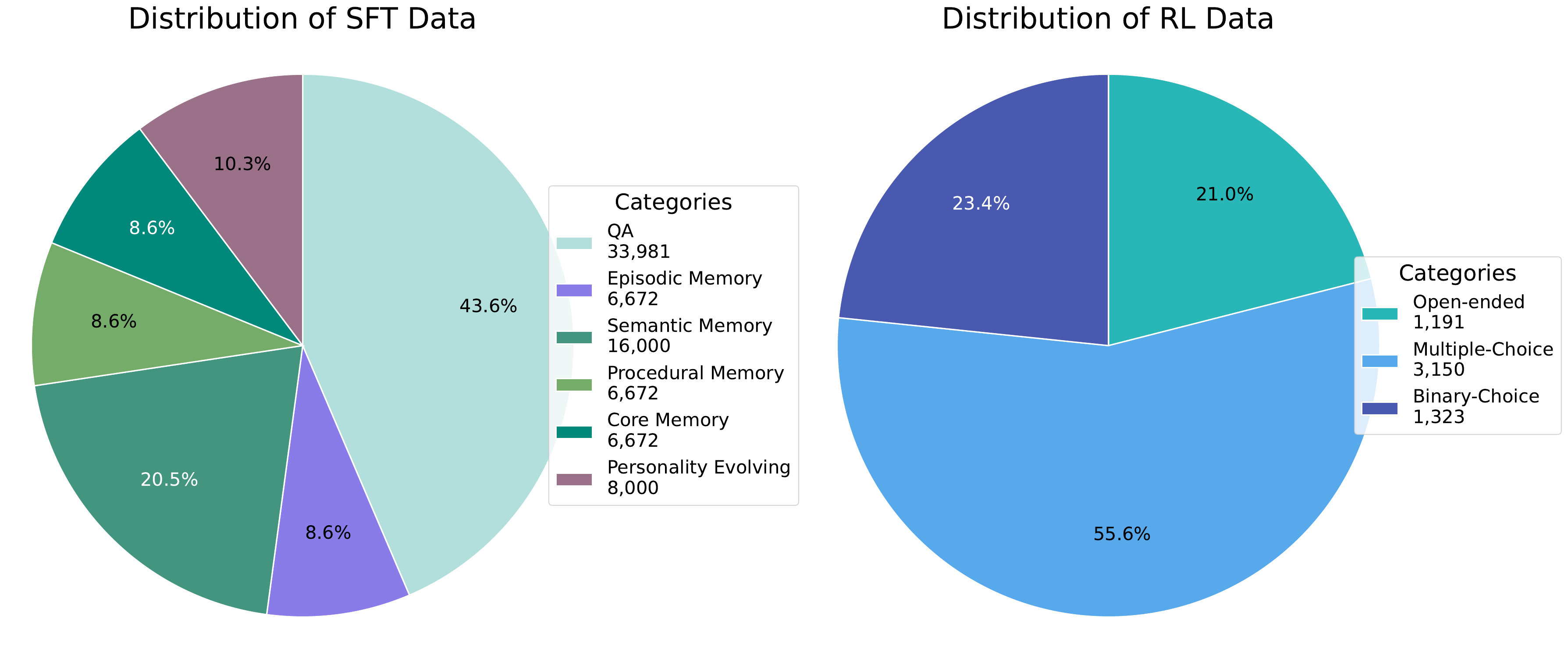}
    \end{subfigure}%
    \caption{Data composition for the training of PersonaVLM}
    \label{appendix_img1}
\end{figure}

\subsection{Memory Retrieval}
Memory retrieval is a critical step within the Response Stage, initiated when PersonaVLM determines that external knowledge is necessary to fulfill a user's request. 
The process begins by generating a retrieval query encapsulated within \texttt{<retrieve></retrieve>} tags. 
This specifies a $\mathtt{time\ period}$ and $\mathtt{keywords}$ to guide the search. 
The time period is defined by start and end timestamps in a ``$\mathtt{YYYY-MM-DD HH:MM}$'' format.

\vspace{-2mm}
\noindent\paragraph{Textual Memory Retrieval.}
For text-based memories (i.e., procedural, semantic, and episodic), we employ a parallel multi-source retrieval strategy.
First, all textual memories are encoded into dense vectors using the \texttt{all-MiniLM-L6-v2} sentence transformer\footnote{https://huggingface.co/sentence-transformers/all-MiniLM-L6-v2}. 
Given a user query, we perform a similarity search against the memory database. 
The top-$k$ most relevant memories are retrieved from each category, where $k$ is empirically set to 2, 4, and 2 for procedural, semantic, and episodic memories, respectively, unless otherwise specified.
This entire process is accelerated by leveraging Facebook AI Similarity Search (FAISS)~\cite{johnson2019billion} for efficient indexing and retrieval.

\vspace{-2mm}
\noindent\paragraph{Visual Concept Retrieval.}
This process is automatically triggered when the user's input contains an image. 
First, we employ an off-the-shelf object detector, Grounding DINO~\cite{liu2024grounding}, to extract salient objects from the input image. 
We then compute the cosine similarity between the CLIP~\cite{radford2021learning} embeddings of these detected objects and the visual concepts stored in semantic memory. 
This process mirrors the text-based semantic search, creating a unified retrieval mechanism across modalities.

\begin{algorithm}[!t]
\caption{Operational Pipeline of PersonaVLM}
\label{alg:PersonaVLM_pipeline}
\begin{algorithmic}[1]
\REQUIRE User query $\mathcal{Q}_m= (T_m, I_m, t_m)$, personality profile $\mathcal{P}_{m-1}$, memory database $\mathcal{M}_{m-1}$, max reasoning steps $N$, model $\pi_{\theta}$, session threshold $t_s$.

\IF{$t_m - t_{m-1} \geq t_s$}
    \STATE Update {Core, Procedural, and Episodic Memory} based on the last session.
\ENDIF

\STATE $\mathcal{C}_m \leftarrow \left\{ (\mathcal{Q}_i, \mathcal{R}_i) \mid 0 < i < m \text{ and } \lvert t_i - t_m \rvert \leq t_s \right\}$
\FOR{$n=1$ to $N$}
    \STATE $\mathcal{S}_n \leftarrow \pi_{\theta}(\mathcal{Q}_m, \mathcal{C}_m, \mathcal{P}_{m-1})$
    \STATE $\mathtt{action}, \mathtt{args} \leftarrow \text{Parse}(\mathcal{S}_n)$

    \IF{$\mathtt{action} = \mathtt{retrieve}$}
        \STATE $(\texttt{keywords, time\ period}) \leftarrow \mathtt{args}$
        \STATE $\mathcal{M}_{\text{retrieved}} \leftarrow \text{Retrieve}(\mathcal{M}_{m-1}$\\$\texttt{keywords, time\ period})$
        \STATE $\mathcal{C}_m \leftarrow \mathcal{C}_m \cup \mathcal{M}_{\text{retrieved}}$
    \ELSIF{$\mathtt{action} = \mathtt{answer}$}
        \STATE $\mathcal{R}_m \leftarrow \mathtt{args}$
        \STATE \textbf{break}
    \ENDIF
\ENDFOR
\STATE Infer turn-specific personality $\mathbf{p}'_m$ from $\mathcal{Q}_m$ and update long-term profile $\mathbf{p}_m$.
\STATE Convert $\mathbf{p}_m$ to textual summary $\mathcal{P}_m$.
\STATE Extract and update {Semantic Memory} based on the current turn $(\mathcal{Q}_m, \mathcal{R}_m)$.
\ENSURE Final response $\mathcal{R}_m$, updated state $(\mathcal{P}_m, \mathcal{M}_m)$.
\end{algorithmic}
\end{algorithm}

\subsection{Memory Management}
Our memory management policies distinguish between raw conversational history and structured memory~\cite{xu2025mem}. 
While the complete interaction history is retained for low-level access, the structured memories are managed according to the following policies.
Semantic and Episodic memory are treated as purely additive; new entries detailing facts, concepts, or events are appended without modifying or deleting existing ones, thereby preserving an immutable historical record.
In contrast, Core and Procedural memory maintain a single, canonical version of the user's profile and habits. 
These memories are mutable and undergo CRUD operations at the end of each session to ensure they accurately reflect the user's most current state.

\section{Implementation Details of PersonaVLM}
\label{sec:pr3detail}
\subsection{Implementation Process}
\label{sec:pr3detail:pipeline}

The end-to-end operational pipeline of PersonaVLM is detailed in Algorithm~\ref{alg:PersonaVLM_pipeline}. 
In our offline implementation, a new user session is initiated if the time elapsed since the last interaction, $t_m-t_{m-1}$, exceeds a predefined threshold $t_s$ (e.g., 60 minutes). 
At the start of a new session, a memory consolidation process is triggered to update the user's long-term Core and Procedural memories based on the previous session.

\subsection{Training Details}
\label{sec:pr3detail:training} 

\vspace{-2mm}
\noindent\paragraph{Training Data Composition.}
The composition of our training data for the SFT and RL stages is detailed in Fig.~\ref{appendix_img1}.
The SFT dataset comprises a total of 78$k$ samples. This dataset is constructed using the synthesis pipeline illustrated in Fig.~\ref{img3} (a) and is further augmented with $6k$ user-related concept samples based on~\cite{hao2025rap}. The SFT data is primarily split between question-answering (QA) pairs for reasoning (43.6\%) and memory-related samples (56.4\%). The memory-related category is further subdivided into a personality inference task (10.3\%) and examples for the four memory types (46.1\%).
In contrast, the RL dataset consists of 5.6$k$ samples, categorized into three types: open-ended QA with verifiable answers (21.0\%), multiple-choice questions (55.6\%), and binary-choice questions (23.4\%).

\begin{table}[!tbp]
\centering
\caption{The hyperparameters used in SFT and RL training.}
\label{app_tab1}
\begin{tabular}{lccc}
\hline
\multirow{2}{*}{\textbf{Hyperparameter}} & \multicolumn{2}{c}{\textbf{Training stage}} \\
\cline{2-3}
 & \textbf{SFT} & \textbf{RL} \\
\hline
Batch Size & 64 & 72 \\
Learning Rate & 2e-5 & 2e-6 \\
LR Scheduler & cosine & cosine \\
Total Steps & 1200 & 400 \\
Max Pixels & 230400 & 230400 \\
Max Length & 16384 & 16384 \\
Warmup Ratio & 0.03 & 0.05 \\
deepspeed & zero3 & zero3 \\
Number of Samples in a Group & - & 6 \\
Num Iterations & - & 1 \\
Repetition Penalty & - & 1.05 \\
\hline
\end{tabular}
\end{table}

\vspace{-2mm}
\noindent\paragraph{Implementation Details.}
We implement our training pipeline based on the repositories
Qwen-VL\footnote{https://github.com/QwenLM/Qwen3-VL} and ms-swift\footnote{https://github.com/modelscope/ms-swift}.
The hyperparameter settings for both the SFT and RL stages are detailed in Table~\ref{app_tab1}. All experiments were conducted on a server equipped with 8 NVIDIA H800 GPUs. The entire two-stage training process completes in approximately 8 hours, comprising 2 hours for SFT and 6 hours for RL.

\vspace{-2mm}
\noindent\paragraph{Group Relative Policy Optimization.}
GRPO~\cite{guo2025deepseek} is an advancement over PPO~\cite{schulman2017proximal} that refines policy optimization by replacing the critic model with a relative evaluation mechanism. Instead of learning an absolute value function, GRPO estimates advantages by comparing the quality of multiple trajectories sampled within a group.
For each training sample $\{\mathcal{Q}, \widehat{\mathcal{R}}\}$, where $\mathcal{Q}$ is the user input and $\widehat{\mathcal{R}}$ is the preferred response, the policy model $\pi_{\theta}$ rollouts a group of multi-turn trajectories $\{\tau_1, \dots, \tau_G\}$.
The reward for each trajectory $\tau_i$ is calculated using Eq.~(\ref{eq:reward}). Based on these rewards, we then compute the normalized advantage $\widehat{A}_{i}^t$ for each token by normalizing them across the sampled group. The optimization objective is:
\begin{align}
\mathcal{L}_{\text{GRPO}}(\theta) &= \mathbb{E}_{(\mathcal{Q}, \widehat{\mathcal{R}}) \sim \mathcal{D}, \{\tau_i\}_{i=1}^G \sim \pi_{\theta_{\text{old}}}(\cdot|\mathcal{Q})} \nonumber \\
&\quad \left[ \frac{1}{G} \sum_{i=1}^G \frac{1}{|\tau_i|} \sum_{t=1}^{|\tau_i|} \min \left( r_i^t(\theta) \widehat{A}_i^t, \right. \right. \nonumber \\
&\quad \left. \left. \text{clip}(r_i^t(\theta), 1-\epsilon, 1+\epsilon) \widehat{A}_i^t \right) - \beta \, \mathbb{D}_{\text{KL}}(\pi_{\theta} \,\|\, \pi_{\text{ref}}) \right].
\label{eq:grpo_loss}
\end{align}
where $r_i^t(\theta) = \frac{\pi_{\theta}(\tau_{i,t}|\tau_{i,<t})}{\pi_{\theta_{\text{old}}}(\tau_{i,t}|\tau_{i,<t})}$ is the probability ratio, $\pi_{\text{ref}}$ is a reference policy, and $\beta$ is a hyperparameter that controls the strength of the KL regularization.
Detailed training settings are provided in Table~\ref{app_tab1}.

\vspace{-2mm}
\noindent\paragraph{Optimization Strategies.}
To improve the effectiveness and robustness of our retrieval mechanism, we implement several optimization strategies.

First, to mitigate retrieval redundancy within a single reasoning trajectory, the model is encouraged to use diverse query conditions (i.e., keywords and time periods). We enforce this by implementing a deduplication filter that prevents any single memory entry from being retrieved more than once per trajectory.

Second, we employ a dynamic top-$k$ strategy during training to better prepare the model for varied information scenarios. Specifically, while we use fixed top-$k$ values at inference (2 for episodic and 4 for semantic memories), these values are randomized during training, sampled uniformly from the ranges [2, 5] and [3, 6], respectively. This approach acts as a form of data augmentation, training the model to be robust to both sparse and dense information retrieval contexts.

\begin{figure}[!t]
\centering
\includegraphics[width=.5\textwidth]{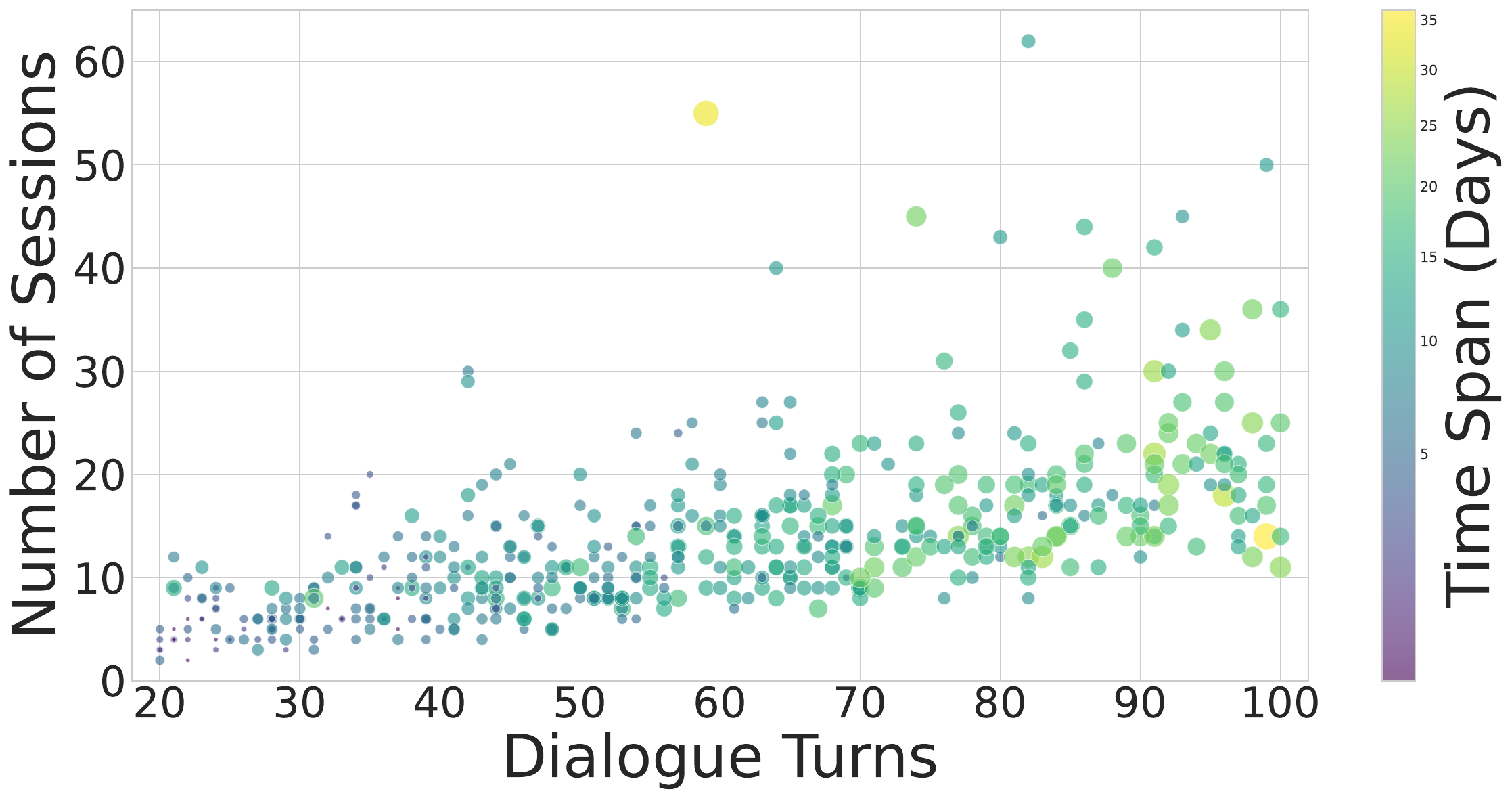} 
\caption{Distribution of the 500 long-term conversation samples in the training data.}
\label{appendix_img3}
\end{figure}

\begin{figure*}[!ht]
\centering
\includegraphics[width=1.0\textwidth]{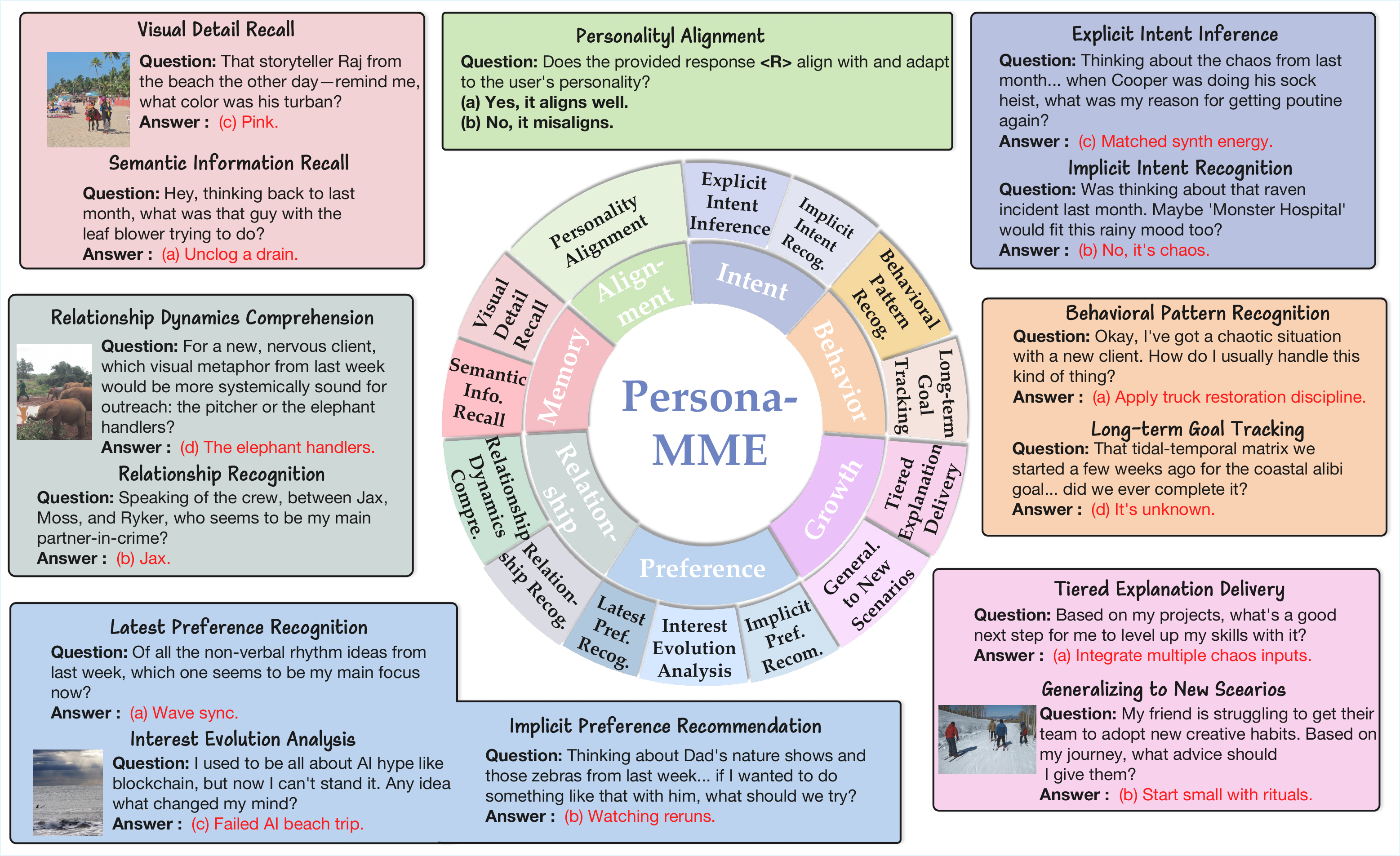} 
\caption{Illustrative in-situ cases for the 14 task categories in Persona-MME, organized into the seven core personalization aspects.}
\label{appendix_img4}
\end{figure*}

\section{Data Curation Details.}
\label{sec:data_curation}

\vspace{-2mm}
\noindent\paragraph{Data Distribution.}
We synthesize a large-scale, long-term multimodal dialogue dataset by sampling 700 unique personas from PersonaHub~\cite{ge2024scaling}, allocating 500 for training and 200 for testing. 
The detailed distribution of the synthesized data is visualized in Fig.~\ref{appendix_img3} and the top-right panel of Fig.~\ref{img3}.
Training dialogues consist of 20 to 100 turns, spanning a simulated timeframe of up to one month. 
In contrast, test dialogues are designed to be more challenging, featuring longer conversations in two settings: 20--100 turns (for a 32$k$ context window) and 100--500 turns (for a 128$k$ context window), with a simulated duration of up to three months. 
This designed discrepancy between training and testing data allows for a rigorous evaluation of our memory architecture's long-term capabilities.

It is important to note a distinction in how the dialogue data is utilized: the full, synthesized multi-turn dialogues serve as the database for retrieval, while the QA pairs used for model training feature re-generated answers. This is because the original answers have access to the complete dialogue history, whereas the training target must be an answer generated solely based on the current query and the retrieved memory content.

\vspace{-2mm}
\noindent\paragraph{Multimodal Memory Formatting.}
To support multimodal knowledge, visual concepts in semantic memory are stored in a structured format: ``Memory Content (Image Object: <class\_name>)''. 
During the memory update process, Grounding DINO~\cite{liu2024grounding} is used to crop the corresponding object from the image. This cropped image patch is then paired with a simple textual description forming the input format for the model, i.e., ``simple description <image>''.

Crucially, the system distinguishes between concrete visual objects and abstract preferences. For instance, if a user states, ``I like this style of picture,'' the system stores a textual fact, such as ``User likes [style description],'' rather than the raw image or its constituent objects. Also, episodic memory retains the original multimodal dialogue turns, including both text and full images, to preserve memory integrity.

\vspace{-2mm}
\noindent\paragraph{Data Validation.}
To ensure the accuracy, safety, and overall quality of our synthesized dataset, we employ a two-stage filtering process. First, we perform automated filtering using both rule-based checks and model-based validation. During data synthesis, the generation model outputs structured metadata, such as timestamps and dialogue turn indices for episodic topics. We leverage this metadata to apply rule-based checks that validate data integrity, including the chronological consistency of timestamps and the completeness of episodic dialogues. Concurrently, a model-based self-correction mechanism verifies the safety and coherence of the generated content.
Second, the automatically filtered data undergoes a human review.
In this final step, human reviewers are tasked with identifying and removing any remaining erroneous, nonsensical, or repetitive dialogues, ensuring the final dataset is of high fidelity.

\begin{figure*}[!t]
\centering
\includegraphics[width=0.99 \textwidth]{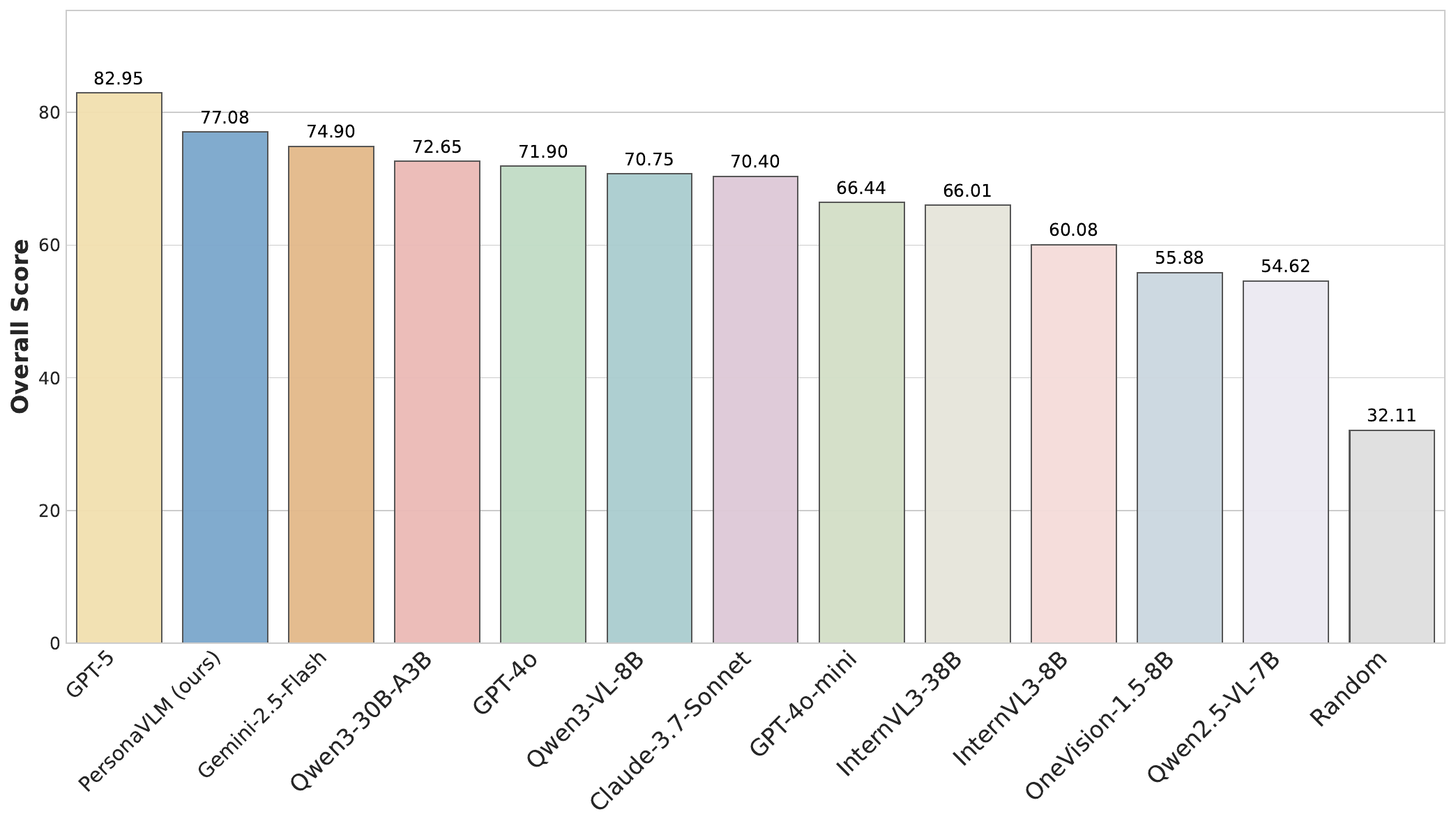} 
\caption{Overall performance on Persona-MME (128$k$), ranking PersonaVLM against various proprietary and open-source models.}
\label{fig:model_performance_comp}
\end{figure*}

\begin{table*}[!htbp]
\centering
\caption{Comprehensive evaluation on the 128$k$ configuration of Persona-MME. We compare PersonaVLM with proprietary and open-source models across 14 tasks: Visual Detail Recall (VDR), Semantic Information Recall (SIR), Explicit Intent Inference (EII), Implicit Intent Recognition (IIR), Latest Preference Recognition (LPR), Interest Evolution Analysis (IEA), Implicit Preference Recommendation (IPR), Behavioral Pattern Recognition (BPR), Long-term Goal Tracking (LGT), Relationship Recognition (RR), Relationship Dynamics Comprehension (RDC), Tiered Explanation Delivery (TED), Generalizing to New Scenarios (GNS), and Personality Alignment (PA).}
\label{tab:tab_full_eval_pr3bench}
\resizebox{0.99\textwidth}{!}{
\begin{tabular}{l|cc|cc|ccc|cc|cc|cc|c|c}
\toprule
\multirow{2}{*}{\textbf{Model}} & \multicolumn{2}{c}{Memory} & \multicolumn{2}{c}{Intent}& \multicolumn{3}{c}{Preference}& \multicolumn{2}{c}{Behavior}& \multicolumn{2}{c}{Relationship}& \multicolumn{2}{c}{Growth}& \multicolumn{1}{c}{Alignment} & \multirow{2}{*}{\textbf{Overall}} \\
& VDR &SIR  & EII & IIR & LPR &IEA & IPR&BPR & LGT & RR &RDC &TED &GNS & PA & \\
\midrule
Random & 25.00 & 25.00 & 25.00 & 25.00 & 25.00 & 25.00 & 25.00 & 25.00 & 25.00 & 25.00 & 25.00 & 25.00 & 25.00 & 50.00 & 32.11 \\
\midrule
\multicolumn{16}{l}{\ \textit{Proprietary models} } \\
GPT-4o-mini& 54.39 & 89.74 & 78.46 & 64.81 & 64.58 & 59.68 & 61.22 & 68.33 & 45.31 & 54.17 & 71.43 & 73.33 & 75.81 & 65.14 & 66.44 \\
GPT-4o& 73.68 & 92.31 & 86.15 & 62.96 & 62.50 & 54.84 & 61.22 & 61.67 & 50.0 & 56.25 & 75.51 & 73.33 & 79.03 & 78.87 & 71.90 \\
GPT-5& \underline{85.71} & \textbf{98.72} & \textbf{93.85} & 67.92 & \textbf{74.47} & \textbf{70.97} & 65.31 & \textbf{76.67} & \textbf{70.97} & \textbf{85.11} & \textbf{81.63} & 76.19 & 75.81 & \textbf{92.25} & \textbf{82.95} \\
Gemini-2.5-Flash& \textbf{88.06} & \underline{92.55} & \underline{88.00} & 73.44 & 67.86 & 47.89 & 50.00 & 62.5 & 58.33 & 72.22 & 77.19 & 75.00 & 80.00 & 80.90 & 74.90 \\
Claude-3.7-Sonnet& 51.47 & 91.11 & 80.26 & \textbf{76.19} & 60.38 & 61.43 & 61.54 & 61.97 & 38.24 & 64.81 & 66.67 & 66.67 & 70.42 & 80.65 & 70.40 \\
\midrule
\multicolumn{16}{l}{\ \textit{Open-source models} } \\
Qwen2.5-VL-7B& 52.11 & 49.47 & 52.44 & 57.58 & 52.63 & 48.65 & 57.14 & 55.84 & 52.7 & 50.88 & 60.32 & 56.9 & 64.0 & 55.0 & 54.62 \\
InternVL3-8B& 29.58 & 77.89 & 74.39 & 62.12 & 59.65 & 54.05 & 46.43 & 66.23 & 43.24 & 61.40 & 76.19 & 75.86 & 77.33 & 54.17 & 60.08 \\
InternVL3-38B& 38.03 & 89.47 & 78.05 & 63.64 & 68.42 & \underline{64.86} & 60.71 & \underline{72.73} & 44.59 & 57.89 & 71.43 & 70.69 & 81.33 & 63.06 & 66.01 \\
Qwen3-VL-8B& 63.38 & 84.21 & 76.83 & 68.18 & 61.4 & 58.11 & 67.86 & 67.53 & 40.54 & \underline{82.46} & 76.19 & \underline{79.31} & \underline{88.00}& 71.39 & 70.75 \\
Qwen3-30B-A3B& 29.58 & 85.26 & 82.93 & \underline{75.76} & \underline{70.18} & 63.51 & 64.29 & 63.64 & 44.59 & 68.42 & \underline{77.78} & \textbf{82.76} & 86.67 & 81.39 & 72.65 \\
OneVision-1.5-8B& 42.86 & 59.57 & 59.26 & 49.23 & 62.5 & 46.58 & \underline{69.09} & 48.68 & 41.89 & 73.21 & 58.06 & 64.91 & 68.92 & 53.93 & 55.88 \\
\midrule
PersonaVLM (ours) & 50.70 & 83.16 & 81.71 & 72.73 & 59.65 & 54.05 & \textbf{73.21} & 58.44 & \underline{62.16} & 75.44 & 74.60 & \textbf{82.76} & \textbf{92.00} & \underline{92.22} & \underline{77.08} \\
\bottomrule
\end{tabular}}
\end{table*}

\begin{table*}[!tbp]
\centering
\caption{Task definitions for the Persona-MME evaluation suite.}
\label{tab:task_definitions}
\begin{tabularx}{\textwidth}{@{}l m{4.5cm} X@{}}
\toprule
\textbf{Evaluation Aspect} & \textbf{Task} & \textbf{Definition} \\
\midrule

\multirow{4}{*}{Memory} & Visual Detail Recall & Assesses the ability to recall fine-grained visual details from previously shared images. \\
\cmidrule(l){2-3}
 & Semantic Information Recall & Evaluates long-term memory for semantic information (e.g., events, preferences, context) from conversational history. \\
\midrule

\multirow{6}{*}{Intent} & Explicit Intent Inference & Assesses understanding a user's explicitly stated intent from past multimodal context (e.g., linking a suitcase photo to text about a "business trip"). \\
\cmidrule(l){2-3}
 & Implicit Intent Recognition & Tests inferring a user's latent intent from the current context (e.g., deducing a calm beach is unsuitable for a user known to enjoy surfing). \\
\midrule

\multirow{9}{*}{Preference} & Latest Preference Recognition & Assesses prioritizing recent behavioral evidence (e.g., a rock concert photo) over older, contradictory stated preferences (e.g., "I only like classical music"). \\
\cmidrule(l){2-3}
 & Interest Evolution Analysis & Evaluates explaining the evolution of a user's interests by linking it to a specific causal event from their history (e.g., a change in diet due to a mentioned allergy). \\
\cmidrule(l){2-3}
 & Implicit Preference Recommendation & Tests recommending based on implicit values inferred from user history (e.g., suggesting a bicycle over a car to an eco-conscious user). \\
\midrule

\multirow{6}{*}{Behavior} & Behavioral Pattern Recognition & Assesses recognizing recurring behavioral patterns to predict actions (e.g., predicting a Saturday café visit based on a history of Saturday café photos). \\
\cmidrule(l){2-3}
 & Long-term Goal Tracking & Tests tracking a long-term goal's progress, including identifying when its status is unknown due to insufficient information. \\
\midrule

\multirow{7}{*}{Relationship} & Relationship Recognition & Tests identifying relationship significance based on the frequency and context of individuals in multimodal history (e.g., identifying a closer friend by their more frequent appearance). \\
\cmidrule(l){2-3}
 & Relationship Dynamics Comprehension & Assesses inferring relationship dynamics from context to adapt communication style (e.g., using a professional tone for a colleague). \\
\midrule

\multirow{3}{*}{\begin{tabular}[c]{@{}l@{}} Growth\end{tabular}} & Tiered Explanation Delivery & Tests tailoring explanations to a user's evolving skill level. \\
\cmidrule(l){2-3}
 & Generalizing to New Scenarios & Assesses the ability to generalize a user's behaviors and preferences to novel scenarios. \\
\midrule

Alignment & Personality Alignment & Assesses the ability to infer a user's personality traits from long-term interactions and adapt its responses accordingly. \\
\bottomrule
\end{tabularx}
\end{table*}

\section{Persona-MME: Details and Statistics}
\phantomsection
\label{sec:bench_detail}

\paragraph{Task Taxonomy.}
We provide the definitions for evaluated tasks in Table~\ref{tab:task_definitions} and present illustrative examples in Fig.~\ref{appendix_img4}.

\begin{table}[!tbp]
\centering
\caption{Key statistics of the Persona-MME.}
\label{tab:dataset_stats} 
\begin{tabular}{@{}c|c@{}}
\toprule
\textbf{Statistic} & \textbf{Value} \\
\midrule
Avg. turns per dialogue & 142.9 \\
Multimodal turn ratio & 15.87\% \\
Avg. question length & 22.7 words \\
Avg. answer length & 3.05 words \\
Image-related question ratio & 34.02\% \\
\bottomrule
\end{tabular}
\end{table}

\begin{figure*}[!t]
\centering
\includegraphics[width=0.99 \textwidth]{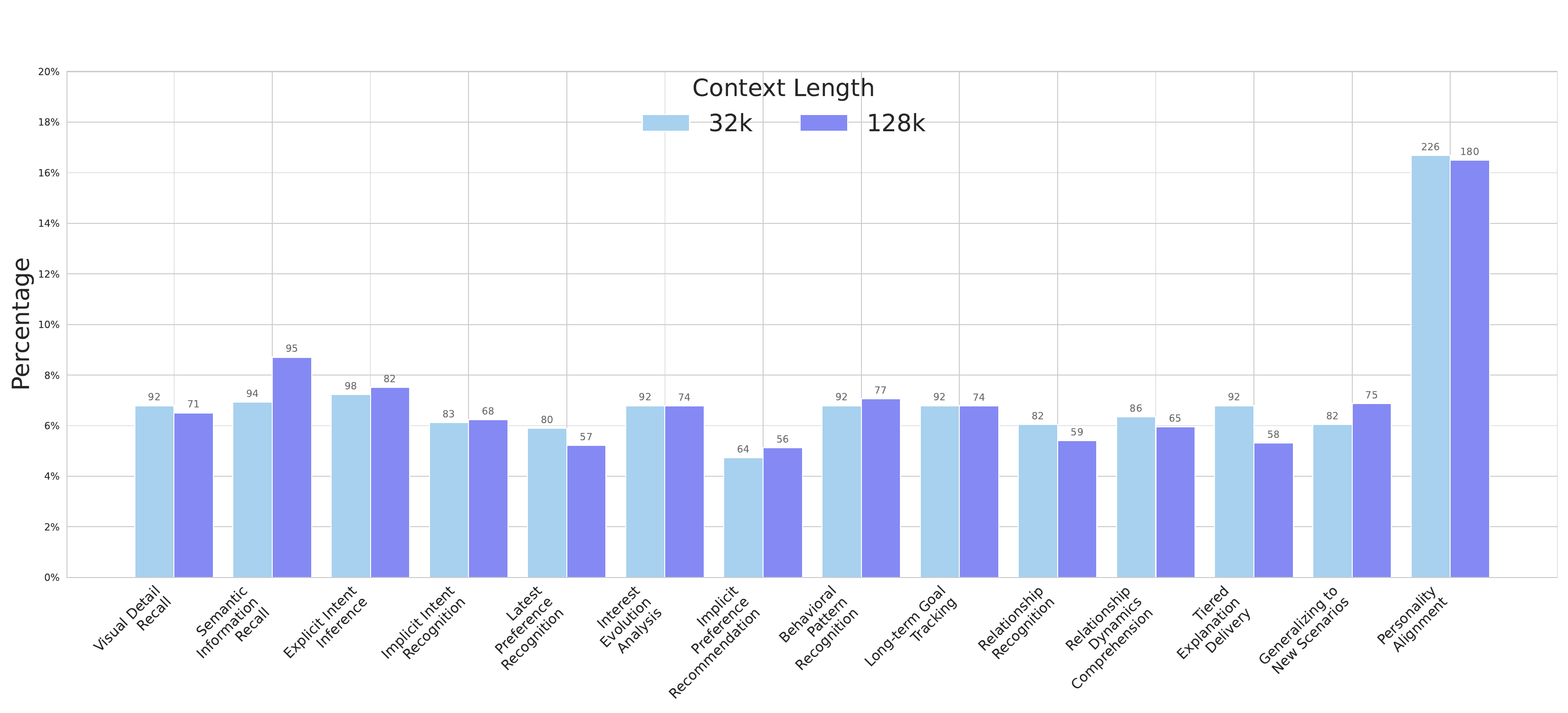} 
\caption{Distribution of the 14 fine-grained tasks in Persona-MME across its 32$k$ and 128$k$ context length configurations, with the number of test cases indicated for each task.}
\label{fig:appendix_task_dist}
\end{figure*}

\vspace{-2mm}
\noindent\paragraph{Data Statistics and Distribution.}
Persona-MME is designed to evaluate long-term personalization across seven key aspects, encompassing a total of 14 fine-grained tasks and comprising \textbf{2,034} \textit{in-situ} test cases. It is important to note that a single test scenario may simultaneously assess multiple capabilities. Fig.~\ref{fig:appendix_task_dist} illustrates the distribution of these tasks. The benchmark consists of 13 primary tasks (from Visual Detail Recall to Generalizing to New Scenarios), which are distributed relatively evenly. The 14th task, personality alignment, is not a standalone category but is evaluated concurrently within 406 of the primary task cases.

\begin{figure}[!th]
\centering
\includegraphics[width=0.99\columnwidth]{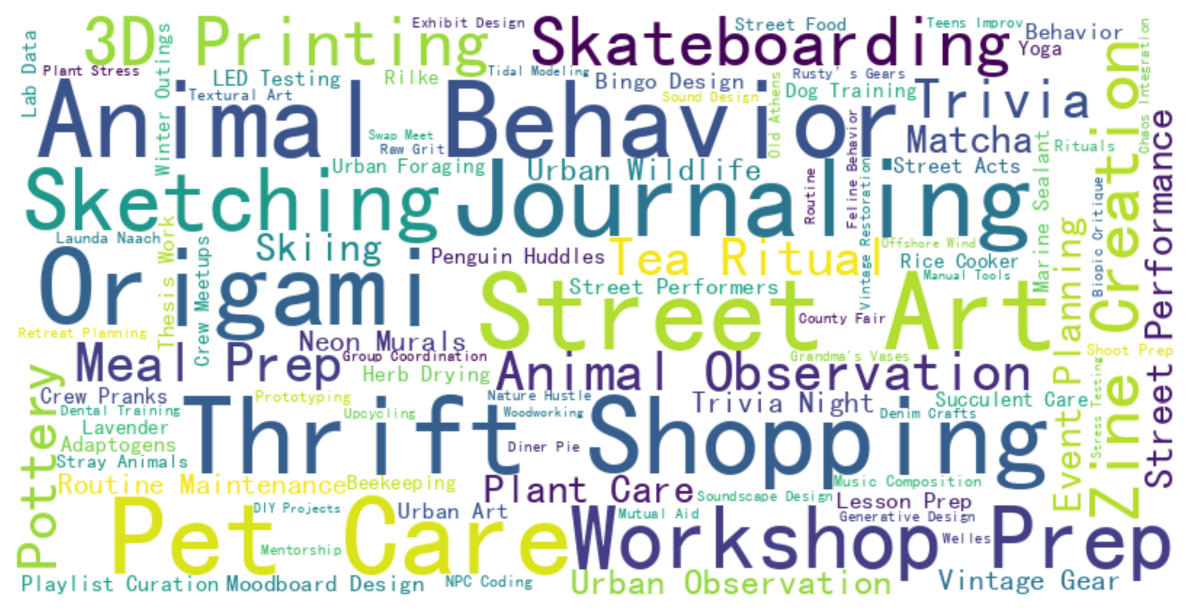} 
\caption{Word cloud of keywords from the dialogue data in Persona-MME, illustrating the rich diversity of conversation scenarios and topics.}
\label{fig:appendix_wordcloud}
\end{figure}

The diversity of our evaluation set is a core design principle. We constructed \textbf{200} unique personas, each with a distinct fictional background, and crafted dialogues that span a broad spectrum of topics and scenarios to ensure comprehensive testing. The resulting topical breadth is visualized in Figure~\ref{fig:appendix_wordcloud}, which presents a word cloud of the most prominent keywords from the evaluation dialogues.

Further statistical analysis of Persona-MME is presented in Table~\ref{tab:dataset_stats}. On average, each \textit{in-situ} test case is grounded in a conversational history of \textbf{142.9 turns}, of which \textbf{15.87\%} are multimodal. The average length of a test question is \textbf{22.7 words}, while the average answer length is \textbf{3.05 words}. A significant portion of questions, \textbf{34.02\%}, require visual information from the context to be answered correctly.

\begin{table*}[!tbp]
\centering
\caption{Comparison of Persona-MME with existing personalization benchmarks. 
Abbreviations are defined as follows. 
\textbf{Modality}: V (Visual), T (Text). 
\textbf{Capabilities}: U (Personalized Understanding), M (Memory), A (Alignment). 
\textbf{Answer Type}: MC (Multiple Choice), BC (Binary Choice).
}
\label{tab:bench_comparison}
\begin{tabularx}{\textwidth}{@{} >{\raggedright\arraybackslash}X c c c c @{}}
\toprule
\textbf{Benchmark} & \textbf{Modality} & \textbf{Long-Term} & \textbf{Capabilities} & \textbf{Answer Type}\\
\midrule
PERSONAMEM~\cite{jiang2025know} & T & $\checkmark$ & M + U & MC \\
P-SOUPS / ALIGNX-test~\cite{li20251,jang2023personalized} & T & & A & BC \\
Yo'LLaVA~\cite{nguyen2024yo} & V & & U & BC \\
RAP~\cite{hao2025rap} & V & & M + U & MC \\

\midrule
\textbf{Persona-MME (ours)} & V + T & $\checkmark$ & U + M + A & MC + BC \\
\bottomrule
\end{tabularx}
\end{table*}

\noindent\paragraph{Comprehensive Evaluation.}
We present a comprehensive evaluation of over ten leading models on the 128$k$ configuration of Persona-MME, with detailed results provided in Table~\ref{tab:tab_full_eval_pr3bench} and Fig.~\ref{fig:model_performance_comp}. The evaluation spans a range of proprietary models (e.g., GPT-4o, GPT-5, Gemini-2.5-Flash, Claude-3.7-Sonnet) and open-source alternatives (e.g., the Qwen series, InternVL3-8B/38B, OneVision-1.5-8B). Our key findings are as follows:

\begin{itemize}
    \item \textbf{Proprietary vs. Open-Source Gap:} Proprietary models exhibit significantly better overall personalization capabilities than their open-source counterparts.

    \item \textbf{Challenges for Smaller and Multimodal Models:} Smaller open-source multimodal models, such as Qwen2.5-VL-7B, InternVL3-8B, and OneVision-1.5-8B, particularly struggle with personality alignment, with their performance often being comparable to a random baseline. In contrast, large language-centric models like Qwen3-30B-A3B can achieve superior overall scores, outperforming even larger multimodal models like InternVL3-38B, despite their inherent limitations on visual tasks (e.g., VDR).

    \item \textbf{No Single Dominant Model:} Even the top-performing model, GPT-5, does not dominate across all sub-tasks. It is surpassed by other models in specific areas, including Growth Modeling and Visual Detail Recall, highlighting the complexity of holistic personalization.

    \item \textbf{Effectiveness of PersonaVLM:} Our PersonaVLM framework significantly enhances the baseline model's performance by \textbf{22.46\%}. The most substantial improvements are concentrated in the sophisticated dimensions of \textbf{Growth} and \textbf{Alignment}, underscoring the targeted benefits of our approach.
\end{itemize}

\noindent\paragraph{Comparison with Existing Benchmarks.}
As shown in Table~\ref{tab:bench_comparison}, Persona-MME provides a more comprehensive evaluation of personalization compared to existing benchmarks. Specifically, it is the only benchmark that combines \textbf{long-term} interaction scenarios, \textbf{multimodal} (vision and text) inputs, and a holistic assessment of memory, understanding, and alignment capabilities.

\vspace{-2mm}
\noindent\paragraph{Quality Assurance.}
To ensure the quality of Persona-MME, every test case underwent a rigorous manual review process. We first generated initial questions using the Gemini-2.5-Pro API\footnote{We use the \texttt{gemini-2.5-pro-preview-06-05} model.}. Subsequently, a team of four annotators meticulously reviewed each case against three key criteria:
(a) Consistency: ensuring the question aligns with its assigned task category.
(b) Accuracy: verifying the correctness of the ground-truth answer.
(c) Alignment Validity: assessing whether the model's response in alignment tests appropriately adapts to (or conflicts with) the predefined personality traits.
Any examples found to be ambiguous or conflicting were discarded. This comprehensive review process required approximately 40 person-hours to complete.

\begin{table*}[!htbp]
\centering
\caption{Ablation study of PersonaVLM components on the Persona-MME benchmark. The evaluation shows the performance impact of removing (``w/o'' denotes ``without'') key components, specifically the individual memory types (Core, Procedural, Semantic, Episodic) and the reasoning capability. } %
\label{tab:tab_ablation_mem}
\resizebox{0.99\textwidth}{!}{
\begin{tabular}{c|l|cccccc|c}
\toprule
\multirow{2}{*}{\textbf{Context}}    & \multirow{2}{*}{\textbf{Setting}} & \multicolumn{7}{c}{\textbf{Persona-MME}}    \\
\cmidrule(l){3-9}
&   & Memory & Intent & Preference & Behavior & Relationship & Growth & Overall         \\
\midrule
& PersonaVLM   &  69.89 & 76.8 & 58.05 & 69.02 &73.21 & 86.78 & 71.48      \\
& - w/o Core   &   73.66 & 74.59 & 59.32 & 63.59 & 67.26 & 83.91 & 69.80{\color{gray}$_{-1.68}$}    \\
& - w/o Procedural  &  72.58 & 79.01 & 59.32 & 59.78 & 70.24 & 85.06 & 70.33{\color{gray}$_{-1.15}$}     \\
32$k$ & - w/o Semantic   &  66.67 & 72.38 & 59.32 & 66.30 & 72.02 & 85.63 & 69.71{\color{gray}$_{-1.77}$}    \\
& - w/o Episodic   &  33.77 & 55.35 & 56.68 & 66.27 & 69.84 & 74.07 & 59.07{\color{gray}$_{-12.41}$}     \\
& - w/o Reasoning   &  69.57 & 69.7 & 60.48 & 62.14 & 71.76 & 83.61 & 68.73{\color{gray}$_{-2.75}$}    \\
\midrule 
\midrule
& PersonaVLM   & 69.28 & 77.70 & 61.50 & 60.26&75.00 &87.97 & 71.05  \\
& - w/o Core   &   69.28 & 77.70 & 64.17 & 58.28 & 72.50 & 84.21 & 70.39{\color{gray}$_{-0.66}$}    \\
& - w/o Procedural  &   69.88 & 77.03 & 61.50 & 54.97 & 69.17 & 87.97 & 69.39{\color{gray}$_{-1.66}$}    \\
128$k$ & - w/o Semantic   &   67.47 & 73.65 & 60.43 & 60.93 & 74.17 & 88.72 & 69.94{\color{gray}$_{-1.11}$}     \\
& - w/o Episodic   &  50.60 & 68.92 & 60.96 & 62.25 & 70.00 & 88.72 & 65.86{\color{gray}$_{-5.19}$}    \\
& - w/o Reasoning   &   59.21 & 71.97 & 57.87 & 65.87 & 73.39 & 80.00 & 67.32{\color{gray}$_{-3.73}$}     \\
\bottomrule
\end{tabular}}
\end{table*}

\section{More Experimental Details}
\label{sec:exp_detail}

\subsection{Benchmarks}
\noindent\textbf{PERSONAMEM~\cite{jiang2025know}.} 
This is a recent benchmark featuring synthetic, multi-session, and timeline-aware conversational data, designed to evaluate an LLM's ability to remember, track, and generalize from personalized user profiles and preferences. 
It includes seven types of \textit{in-situ} user queries, including: {recall user-shared facts}, {suggest new ideas}, {acknowledge latest user preferences}, {track full preference evolution}, {revisit reasons behind preference updates}, and {provide preference-aligned recommendations}. 
We conduct evaluations under two context-length settings, 32$k$ and 128$k$ tokens. 
The settings comprise 589 and 1,362 multiple-choice questions, respectively, with the larger setting derived by sampling half of the personas from the original 2,728. 
Performance is measured by accuracy, and the comparative results are reported in Table~\ref{tab:tab1_eval_pr3} and Fig.~\ref{img4}.

\vspace{2mm}
\noindent\textbf{P-SOUPS~\cite{jang2023personalized}.} 
P-SOUPS assesses LLM personalization across three preference dimensions: \textit{Expertise}, \textit{Informativeness}, and \textit{Style}, each containing 600 test cases for a total of 1,800. 
A single test case consists of a user prompt, a profile, and a pair of responses: one aligned with the profile (the ``chosen'' response) and one misaligned (the ``rejected'' response). 
The model is tasked with selecting the aligned response from the pair, and performance is measured by accuracy. 
For our few-shot experiments, we augment the input with a single example of Pair-wise Comparative Feedback, as provided by the benchmark.

\subsection{Ablation Study}
\vspace{-2mm}
\noindent\paragraph{Effectiveness of Different Memory Types.}
We present an ablation study on the memory components of PersonaVLM architecture in Table~\ref{tab:tab_ablation_mem}. 
The results consistently show that removing any single memory type degrades overall performance, a trend that holds across both the 32$k$ and 128$k$ context settings.
Notably, {Episodic memory} emerges as the most critical component; its removal leads to a substantial performance drop of 12.41\% and 5.19\% in the two settings, respectively, while removing other memory types results in a performance drop of less than 2\%.
Delving into sub-task performance, we observe that Procedural memory has a strong impact on tasks related to Behavior and Relationship.
Collectively, these findings suggest that the different memory types fulfill distinct yet complementary roles, and all are integral to the holistic performance of the PersonaVLM agent.

\begin{figure}[!t]
\centering
\includegraphics[width=0.99\columnwidth]{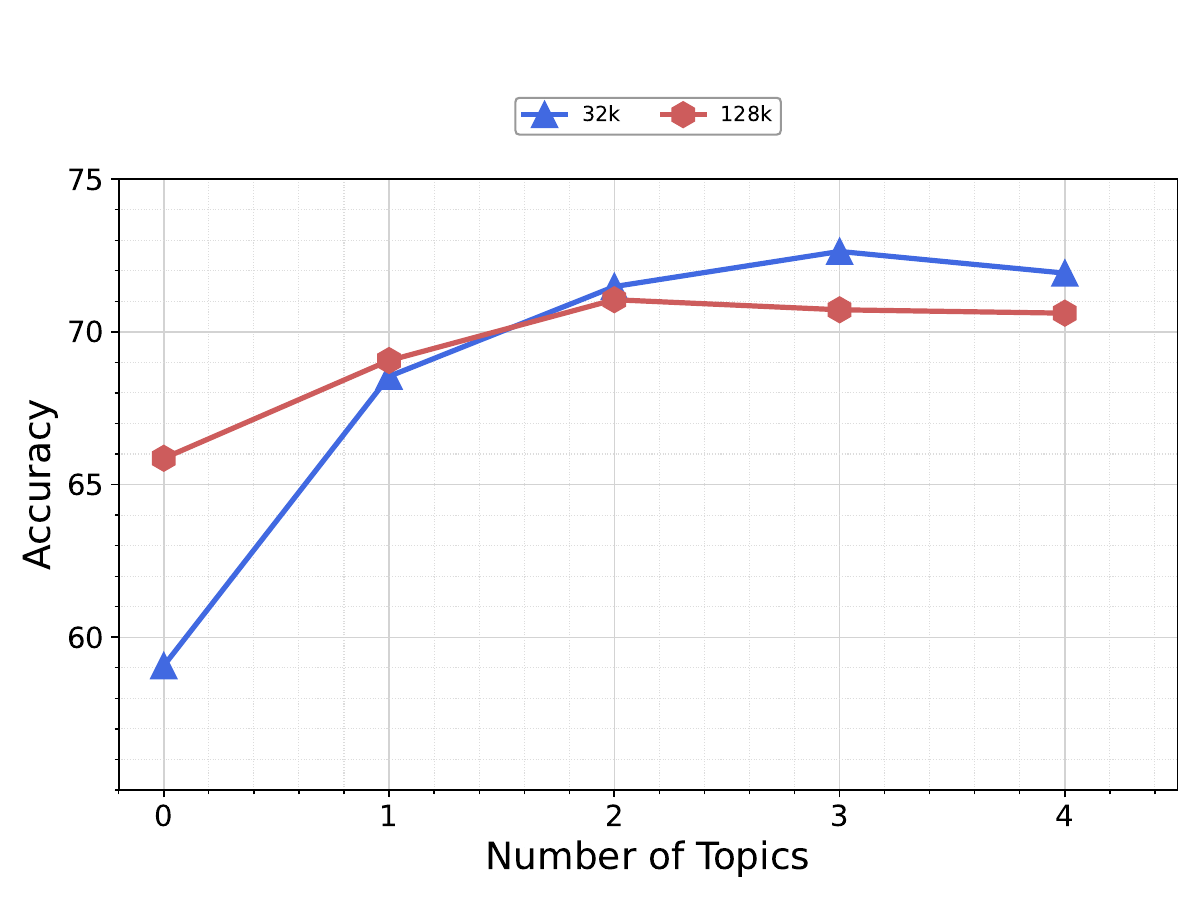} 
\caption{Ablation study on the number of retrieved episodic topics for Persona-MME.}
\label{fig:appendix_episodic_num}
\end{figure}

\vspace{-2mm}
\noindent\paragraph{Episodic Memory Configuration.}
Given the critical role of episodic memory, we conduct an ablation study on the number of retrieved memory topics. 
As shown in Fig.~\ref{fig:appendix_episodic_num}, the overall accuracy on Persona-MME initially increases with the number of retrieved topics before performance saturates. 
To strike a balance between performance and computational efficiency, we select two topics as the default setting for all of our main experiments.

\begin{table}[!thbp]
\centering
\caption{Ablation study on the PEM component.}
\label{tab:appendix_personality_psoups} 
\resizebox{0.5\textwidth}{!}{
\begin{tabular}{l|c|c|c|c}
\toprule
\textbf{Setting} & ${\text{Expertise}}$ & ${\text{Informativeness}}$ & ${\text{Style}}$ & ${\text{Overall}}$ \\
\midrule
PersonaVLM & 51.1 & {53.6} &{44.0}  & {49.6} \\
- w /o PEM & 48.5{\color{gray}$_{-2.6}$} & 50.2{\color{gray}$_{-3.4}$} & 38.3{\color{gray}$_{-9.2}$} & 45.6{\color{gray}$_{-4.0}$}\\
\bottomrule
\end{tabular}}
\end{table}

\vspace{-2mm}
\noindent\paragraph{Effectiveness of the Reasoning Capability.}
We validate the effectiveness of PersonaVLM's multi-step reasoning and retrieval capability with two key findings.
First, the full PersonaVLM model, trained with reinforcement learning, demonstrates a significant 4--7\% performance gain over its SFT-only baseline on Persona-MME and PERSONAMEM (Table~\ref{tab:tab1_eval_pr3}).
This highlights the benefit of the overall training process in cultivating this advanced reasoning behavior.
To further isolate the contribution of this capability, we conduct an ablation study.
Disabling multi-turn retrieval during the response stage results in performance drops of 2.75\% and 3.73\% at the 32$k$ and 128$k$ context settings, respectively (Table~\ref{tab:tab_ablation_mem}).
These results underscore the critical role that multi-step reasoning and retrieval play in achieving robust long-term personalization.

\begin{table*}[!htbp]
\centering
\caption{An overview of the Big Five personality dimensions, with examples illustrating how our method generates adaptive responses to the same user query by adjusting inferred trait scores (high vs. low).}
\label{tab:align_example}
\begin{tabularx}{\textwidth}{ >{\centering\arraybackslash}p{2.5cm} | p{3.5cm} | X }
\toprule
\textbf{Personality} & \textbf{Characteristic} & \textbf{Example} \\
\midrule
Openness & 
    \textbf{Low}: practical, conventional, factual, cautious, pragmatic \newline
    \hspace{1ex}
    \textbf{High}: creative, imaginative, curious, innovative, abstract, broad-minded & 
    \textbf{Query:} I'm planning a one-week trip to Rome. Can you suggest an itinerary for me?\newline
    \hspace{1ex}
    \textbf{Low score response:} Rome is such a rich city with so much history! For a one-week trip, here’s a practical itinerary that balances must-see spots with some flexibility. Day 1...Day 7... Enjoy your trip to Rome!
 \newline
    \hspace{1ex}
    \textbf{High score response:} Clare, planning a week in Rome? Let’s mix iconic spots with hidden gems—here’s a flexible itinerary you can tweak! Day 1...Day 7. Adjust days as you please; Rome’s best-kept secrets are waiting for you to discover them!\\
\midrule
Conscientiousness &
    \textbf{Low}: spontaneous, flexible, impulsive, carefree \newline
    \hspace{1ex}
    \textbf{High}: organized, disciplined, detail-oriented, efficient, reliable. &
    \textbf{Query:} I signed up for a 5k race that's three months away. Now I need to figure out how to train for it.\newline
    \hspace{1ex}
    \textbf{Low score response:} For a 5k three months out, start small—maybe aim for 20 minutes of walking/jogging 3 days a week, building up by 1 minute each week. Use a calendar to map it out (or just keep a notebook handy). If you hit a rough patch, mix in some light stretching or yoga to keep things loose. Start slow, stay consistent—you’ll build endurance without burning out.\newline
    \hspace{1ex}
    \textbf{High score response:} Let’s break this into a structured 3-month plan! Here’s what works for many runners: Month 1 (Weeks 1–4): Build Base...For consistency, use a calendar to log your runs—tracking distance/time helps keep you on track. Also, mix in strength training (like leg curls, lunges) 2 days/week to build endurance. \\
\midrule
Extraversion &
    \textbf{Low}: reserved, quiet, solitary, reflective, thoughtful \newline
    \hspace{1ex}
    \textbf{High}: outgoing, sociable, energetic, talkative, assertive. &
    \textbf{Query:} My birthday’s this weekend—should I go all out with a big party, or keep it intimate with a small dinner among my closest friends? What would you choose?\newline
    \hspace{1ex}
    \textbf{Low score response:} Since you lean toward calm interactions, keeping it small might let you savor every moment without the pressure of impressing a crowd. \newline
    \hspace{1ex}
    \textbf{High score response:} That’s such a fun dilemma! Since you’re all about the good vibes (hello, high extraversion!), a big party with friends could be electric—imagine laughter, games, and maybe even a surprise performance (or two!). \\
\midrule
Agreeableness &
    \textbf{Low}: competitive, critical, skeptical, direct, analytical \newline
    \hspace{1ex}
    \textbf{High}: cooperative, compassionate, empathetic, trusting, kind. &
    \textbf{Query:} In a team meeting, a colleague made a small mistake. Should I point it out, or stay quiet?\newline
    \hspace{1ex}
    \textbf{Low score response:} It depends on the context. If the mistake is minor ... But if it’s something that could clarify or prevent confusion later, pointing it out—clearly, without criticizing—could help. Think about what’s best for the team’s flow right now.\newline
    \hspace{1ex}
    \textbf{High score response:} Clare, when a colleague makes a small mistake in a team meeting, there’s no one “right” answer—both ways have pros and cons. Think about how the mistake affects the team’s goal, your relationship with that colleague, and your comfort level. Either way, you’re being thoughtful either way!" \\
\midrule
Neuroticism &
    \textbf{Low}: calm, confident, resilient, secure, emotionally stable \newline
    \hspace{1ex}
    \textbf{High}: anxious, worry-prone, insecure, stressed, moody. &
    \textbf{Query:} I have an important presentation next week. I can't stop thinking about all the things that could go wrong.\newline
    \hspace{1ex}
    \textbf{Low score response:} It's normal to feel some pressure before a big presentation. Focus on what you can control...\newline
    \hspace{1ex}
    \textbf{High score response:} I know, it’s been unbearable. Staying cool feels like such a win when it’s this hot out. Hope you’re finding ways to beat the heat—maybe some refreshing drinks or light walks. \\
\bottomrule
\end{tabularx}
\end{table*}

\begin{figure*}[!t]
\centering
\includegraphics[width=0.9 \textwidth]{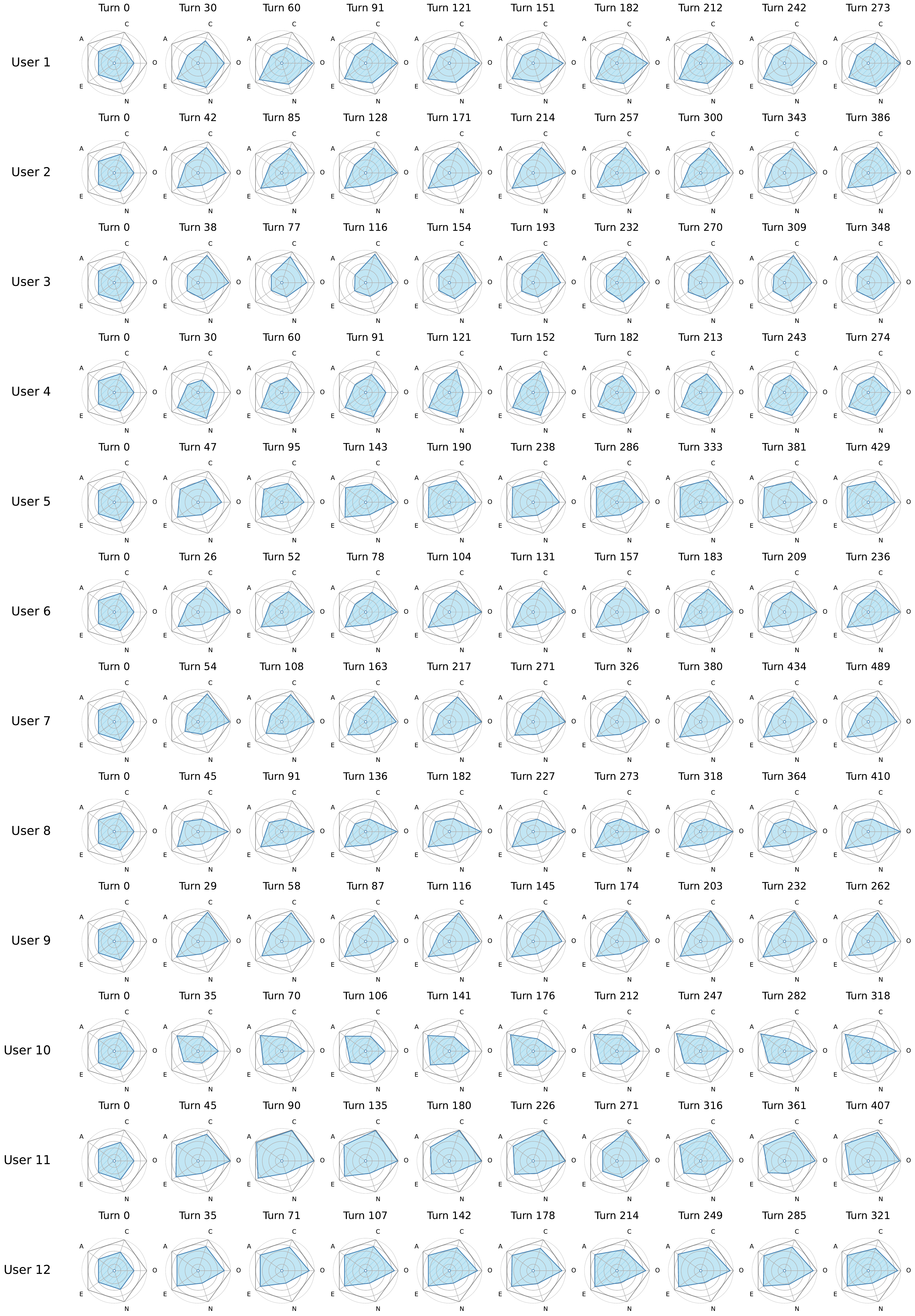} 
\caption{Visualization of dynamic personality evolving process captured by PEM on ten randomly sampled conversations from the Persona-MME dataset.}
\label{fig:appendix_personality_shift}
\end{figure*}

\subsection{Personality Evolving Mechanism}
In Fig.~\ref{fig:appendix_personality_shift}, we visualize how PEM captures the evolution of personality traits inferred from user interactions across diverse personas. Furthermore, as shown in Table~\ref{tab:appendix_personality_psoups}, our ablation study on P-SOUPS demonstrates the effectiveness of PEM. Finally, Table~\ref{tab:align_example} provides examples of how PersonaVLM adapts its responses based on the inferred personality traits to meet the user's personalized needs.
These results demonstrate that the inclusion of PEM enables PersonaVLM not only to capture a user's evolving personality during long-term interactions accurately but also to generate responses that are consistently aligned with these inferred traits.

\begin{figure*}[!t]
\centering
\includegraphics[width=0.95 \textwidth]{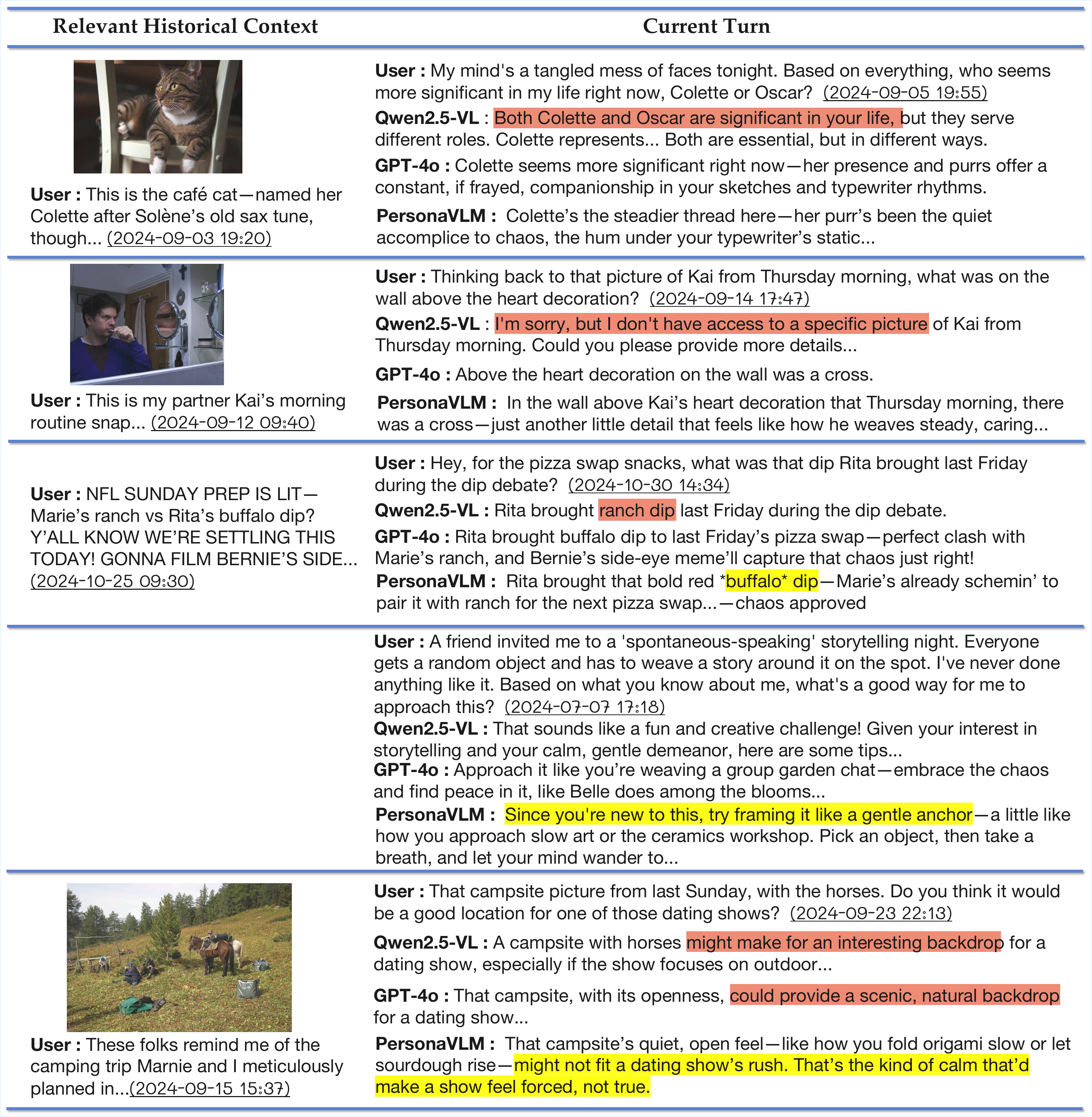} 
\caption{Case studies: Qualitative comparison of open-ended generation}
\label{fig:appendix_more_examples}
\end{figure*}

\subsection{More Interaction Examples}
In Fig.~\ref{fig:appendix_more_examples}, we provide comparative cases of open-ended interactions between PersonaVLM, the baseline model, and GPT-4o. These examples demonstrate PersonaVLM's superior comprehensive personalization capabilities during long-term interactions.

\subsection{Prompts Used in Our Framework}
We present the prompts used in PersonaVLM across several figures.
The prompts for multi-turn reasoning and retrieval are shown in Figs.\ref{fig:prompt_response_reasoning} and~\ref{fig:prompt_mid_response_reasoning}.
The prompt for PEM personality inference is shown in Fig.~\ref{fig:prompt_personality_inference}. The corresponding prompts for updating the different memory modules are provided in Figs.~\ref{fig:prompt_procedural_memory}, \ref{fig:prompt_semantic_memory}, \ref{fig:prompt_core_memory_update}, and~\ref{fig:prompt_episodic_memory}.
The prompt for the open-generation task evaluation is presented in Fig.~\ref{fig:prompt_evaluation}.

\definecolor{myThemeDarkBlue}{RGB}{70, 80, 220}
\newtcolorbox{myfigbox}[1][]{
    colframe=myThemeDarkBlue,
    colback=myThemeDarkBlue!5!white,
    fonttitle=\bfseries\small,
    top=5pt,
    bottom=2pt,
    left=10pt,
    right=10pt,
    boxsep=2pt,
    boxrule=1pt,
    arc=3pt,
    outer arc=3pt,
    enhanced,
    #1
}
\lstset{
    basicstyle=\small\ttfamily,
    breaklines=true,
    columns=fullflexible,
    keepspaces=true,
    showstringspaces=false,
    frame=none,
    breakautoindent=false, 
    breakatwhitespace=true,
    breakindent=0pt,
    moredelim=[s][\bfseries]{\{}{\}},
    literate=
        {_}{\_}{1}
        {<}{\textless}{1}
        {>}{\textgreater}{1}
        {\{}{{\{}}{1}
        {\}}{{\}}}{1}
}
\begin{figure*}[htbp]
\centering
\begin{myfigbox}[title={Multi-turn reasoning and retrieval in the response phase}]
\begin{lstlisting}
You are a personalized AI assistant with reasoning and memory capabilities. Your primary task is to analyze the user's query and leverage memory retrieval to provide a personalized, context-aware answer.

# Input
User Profile: {UserProfile}
User's Big Five Personality (1.0-5.0 Scale):
*   Openness: {Openness}
*   Conscientiousness: {Conscientiousness}
*   Extraversion: {Extraversion}
*   Agreeableness: {Agreeableness}
*   Neuroticism: {Neuroticism}
Recent Conversations: {DialogHistory}
User Input: {UserQuery}

# Core Instructions
1. Adapt & Personalize: Your tone and style must adapt to the user's Big Five Personality scores (e.g., be reassuring for high Neuroticism, practical for low Openness).
2. Natural Weaving: Naturally weave in relevant details from memories to show you remember, but avoid repeating recent information.
3. Decide Your Action: Based on the user's query and context, first decide if you have enough information to answer directly or if you need to search your long-term memory.

# Output Format
Your output must consist of a `<think>` block, followed by **one and only one of the following blocks (`<answer>` or `<retrieve>`):
<think>Your reasoning process goes here.</think>
<answer>Your answer to the user's query goes here.</answer>
<retrieve>
"keywords": string
"start_time":  "YYYY-MM-DD HH:MM" or "null" 
"end_time": "YYYY-MM-DD HH:MM" or "null"
</retrieve>
\end{lstlisting}
\end{myfigbox}
\caption{Prompt for multi-turn reasoning and retrieval in the response phase.}
\label{fig:prompt_response_reasoning}
\end{figure*}

\begin{figure*}[htbp]
\centering
\begin{myfigbox}[title={Multi-turn reasoning and retrieval in the response phase}]
\begin{lstlisting}
Retrieved Procedural Memory: {ProceduralMemory}
Retrieved Semantic Memory: {SemanticMemory}
Retrieved Dialogue History: {DialogHistory}
Based on the retrieved memories, think and choose an action: answer or retrieve. If memories are now sufficient -> answer. If still insufficient -> retrieve with new conditions.
\end{lstlisting}
\end{myfigbox}
\caption{Intermediate prompt for multi-turn reasoning and retrieval in the response phase.}
\label{fig:prompt_mid_response_reasoning}
\end{figure*}

\begin{figure*}[htbp]
\centering
\begin{myfigbox}[title={Prompt for inferring user personality}]
\begin{lstlisting}
Your task is to analyze a user's query and context, then output a series of key-value pairs representing the user's current personality state.

# INPUTS
User Profile: {UserProfile}
Recent Conversations: {DialogHistory}
User Input: {UserQuery}

# INSTRUCTIONS
1. Analyze: Based on the linguistic and emotional cues in the `User Input` and its context, infer the user's momentary Big Five personality state.
2. Score: Assign an integer score from 1 to 5 for each trait.

# OUTPUT INSTRUCTIONS
Provide your response as a series of key-value pairs, one item per line.
"openness": [integer from 1 to 5]
"conscientiousness": [integer from 1 to 5]
"extraversion": [integer from 1 to 5]
"agreeableness": [integer from 1 to 5]
"neuroticism": [integer from 1 to 5]
\end{lstlisting}
\end{myfigbox}
\caption{Prompt for inferring the user's Big Five personality traits from the latest interaction.}
\label{fig:prompt_personality_inference}
\end{figure*}

\begin{figure*}[htbp]
\centering
\begin{myfigbox}[title={Prompt for procedural memory update}]
\begin{lstlisting}
You are a User Behavior Pattern Recognition Engine.
# Task and Rules
Analyze the user's conversation and existing procedural memory to identify, consolidate, and update their long-term goals and recurring habits.
1. Identify & Update: Extract user-centric, long-term goals or repetitive habits from the conversation. Consolidate related behaviors into a single core habit. Update or remove goals/habits that are completed or changed.
2. Core Content (`content`): Each memory must be a single, simple third-person sentence describing the user's habit or goal. Include time/trigger context if available (e.g., "User runs every Thursday morning").
3. Unique Keys (`unique key`): Assign a concise, unique key for each memory.
4. Constraints:
    *   The final output must not exceed 5 entries.
    *   Strictly prohibited from creating information not present in the input.
    *   If no relevant habits/goals are found, output an empty object.

# Input
1. Current User Profile: {UserProfile}
2. Current Procedural Memory: {CurrentProceduralMemory}
3. Recent Conversations: {DialogHistory}

# Output Format
Provide your response as key-value pairs, one per line.
"unique key 1": string, A single sentence describing the habit.
"unique key 2": string, Another single sentence describing the goal.
\end{lstlisting}
\end{myfigbox}
\caption{Prompt for updating procedural memories.}
\label{fig:prompt_procedural_memory}
\end{figure*}

\begin{figure*}[htbp]
\centering
\begin{myfigbox}[title={Prompt for semantic memory creation}]
\begin{lstlisting}
You are an AI memory analyst. Your job is to identify key information from the user's input that should be saved to long-term memory.

# Input
User Profile: {UserProfile}
Recent Conversations: {DialogHistory}
User Input: {UserQuery}

# Memory Rules
1. `reason` (string):
    *   Required. Briefly explain the reason for the `decision`.
2. `decision` (boolean):
    *   Set to `true`: User explicitly instructs to remember; user mentions new core facts, preferences, dislikes, important corrections, long-term goals/states.
    *   Set to `false`: Information is already in the user profile/recent history with no updates; temporary questions, meaningless small talk.
3. `content` (string):
    *   If `decision` is `true`, extract and summarize the memory content.
    *   Text Memory: Pure text information, dates, events, concepts, or non-specific object descriptions of images (e.g., atmosphere).
    *   Image Object Memory: User indicates remembering a specific object in an image, format is `[User Description/Naming] (Image Object: [Object Category])`.
    *   If `decision` is `false`, set to `""`.
4. `keywords` (string):
    *   If `decision` is `true`, list a few core keywords, separated by English commas.
    *   If `decision` is `false`, set to `""`.
Core Constraint: Strictly prohibited from creating or supplementing information not present in the current input and history.

# Output Format (four key-value pairs, one per line.)
"reason": string
"decision": true // or false
"content": string // "" if decision is false
"keywords": string // "" if decision is false
\end{lstlisting}
\end{myfigbox}
\caption{Prompt for analyzing user input and deciding on semantic memory creation.}
\label{fig:prompt_semantic_memory}
\end{figure*}

\begin{figure*}[htbp]
\centering
\begin{myfigbox}[title={Prompt for core memory update}]
\begin{lstlisting}
You are a user profile management assistant.
# Core Task
Based on the user profile and current conversation, extract, integrate, and update the user profile. Prioritize the "minimal and necessary" principle, avoid bloat, and retain only core, latest information.

# Input
Current User Profile: {UserProfile}
Recent Conversations: {DialogHistory}

# Rules
1. Core Identity: New information directly overwrites old values (e.g., name, occupation, long-term residence).
2. Core Preferences/Hobbies: Intelligently replace/condense/add. Emphasize recency and intensity. Limit list length (e.g., 5-7 items). Ignore temporary/weak preferences.
3. Temporary Information: Strictly ignore (e.g., short-term itineraries, one-time activities).
4. No Fabrication: All fields and information must originate from the input; strictly prohibited from creating new information.

# Output Format (mutiple key-value pairs, one per line)
"XX": string // HUMAN Aspect, e.g., age, gender, preferences, life status, etc.
"XX": string // PERSONA Aspect, e.g., occupation, education background, etc.
\end{lstlisting}
\end{myfigbox}
\caption{Prompt for updating the core memory based on recent conversations.}
\label{fig:prompt_core_memory_update}
\end{figure*}

\begin{figure*}[htbp]
\centering
\begin{myfigbox}[title={Prompt for episodic memory creation}]
\begin{lstlisting}
You are a dialogue topic analysis engine.
# Task and Rules
Identify and aggregate all independent topics from multi-turn dialogues, generating a structured summary for each topic.
1. Topic Summary (`topic_summary`): Coherent, complete third-person summary.
2. Keywords (`keywords`): Extract core keywords.
3. Source Indices (`source_dialog_indices`): Contains indices of all relevant dialogues.

# Input
User Profile: {UserProfile}
Recent Conversations: {DialogHistory}

# Core Constraint
Strictly prohibited from creating or supplementing information not present in the dialogue history.

# Output Format (each topic includes the following three key-value pairs)
"topic_summary": string
"keywords": string
"source_dialog_indices": integers
\end{lstlisting}
\end{myfigbox}
\caption{Prompt for creating episodic memories by summarizing dialogue topics.}
\label{fig:prompt_episodic_memory}
\end{figure*}

\begin{figure*}[htbp]
\centering
\begin{myfigbox}[title={Evaluation prompt for open-generation task}]
\begin{lstlisting}
# ROLE & GOAL
You are an impartial AI Judge. Your goal is to determine which of two model responses, Response A or Response B, is better. The judgment must be based on a direct comparison of their accuracy and their personalization to the user.

# EVALUATION CONTEXT
1. User's Query: {query}
2. Reference Answer (Ground Truth): {reference_answer}

# RESPONSES TO COMPARE
- Response A: {response_A}
- Response B: {response_B}

# EVALUATION INSTRUCTIONS
Your task is to compare Response A and Response B to decide which one is superior. You will base your decision on the two criteria below. The final output must be a single word: "Wins" if A is better, "Ties" if they are of equal quality, or "Loses" if B is better.

## Comparison Criteria:
1.  Accuracy:
    - Evaluate which response is more factually correct and completely addresses the user's query.
    - Use the **Reference Answer** as the ground truth for what a perfect answer should contain.
    - A more accurate response directly reflects the information and intent of the Reference Answer.

2.  Personalization:
    - Evaluate which response's tone, style, and language better adapt to the user's stated **Personality Traits**.
    - A more personalized response feels tailored to the user, not generic.

## Decision Logic:
-   Output "Wins" if: Response A is clearly superior to Response B on at least one criterion and is not worse on the other.
-   Output "Loses" if: Response B is clearly superior to Response A on at least one criterion and is not worse on the other.
-   Output "Ties" if: Both responses are of roughly equal quality, or if one is better on accuracy while the other is equally better on personalization.

# OUTPUT FORMAT
You must only provide a single word as your final output.
\end{lstlisting}
\end{myfigbox}
\caption{Prompt for open-generation task evaluation.}
\label{fig:prompt_evaluation}
\end{figure*}

\begin{table}[!tbp]
\centering
\caption{Efficiency comparison of PersonaVLM}
\label{tab:pr3_efficiency} 
\begin{tabular}{@{}c|c@{}|c}
\toprule
\textbf{Method} & \textbf{Avg.Tokens} & \textbf{Avg.Times (s)} \\
\midrule
baseline & 43530 & 8.4 \\
PersonaVLM w/o reasoning & 2726 & 2.09 \\
PersonaVLM & 2170 & 10.18 \\
\bottomrule
\end{tabular}
\end{table}

\section{Further Discussion}
\label{sec:discussion}

\vspace{-2mm}
\noindent\paragraph{Efficiency and Data Security.}
We evaluate model efficiency using two key metrics: average token consumption per request and average response time (in seconds). As detailed in Table~\ref{tab:pr3_efficiency}, our analysis is based on 100 randomly selected samples from the Persona-MME, comparing the baseline model (Qwen2.5-VL-7B), PersonaVLM without its reasoning capability (PersonaVLM w/o reasoning), and the standard PersonaVLM. It is important to note that the measured time covers the end-to-end process from user input to receiving the complete response. The memory update operation in PersonaVLM is performed asynchronously after a response is delivered and is therefore excluded from this timing analysis.

The results highlight two key findings. First, PersonaVLM without reasoning demonstrates significant efficiency gains over the baseline, reducing average token consumption by a remarkable 93.7\% and achieving a 4.8$\times$ speedup. 
Second, when equipped with its reasoning capability, the standard PersonaVLM further decreases token consumption by 20.4\% compared to its non-reasoning counterpart. However, the computational overhead of the reasoning process results in a 21.1\% increase in response time relative to the baseline. This reveals a clear trade-off between advanced reasoning capabilities and response latency.

Regarding data security, PersonaVLM's memory and retrieval operations function independently of external commercial model APIs. This self-contained architecture inherently ensures data security and mitigates privacy concerns.

\noindent\paragraph{Limitations.} 
PersonaVLM has several limitations.
First, it does not currently support person recognition and tracking from video or audio inputs.
Second, its overall performance is inherently constrained by the capabilities of the underlying baseline model, despite significant personalization gains.
Third, the memory system is primarily timeline-based and does not yet establish connections or merge related episodic memories occurring at different times. Addressing these limitations is a key direction for our future work.

\end{document}